\documentclass[AMA,STIX1COL]{WileyNJD-v2}

\articletype{Original Articles}%

\received{26 April 2022}
\revised{6 June 2022}
\accepted{6 June 2022}

\usepackage{amsmath,amsthm,amssymb,amsfonts,mathtools}
\usepackage{graphicx}
\usepackage{float}
\usepackage{url}
\usepackage{textcomp}
\usepackage{caption}
\usepackage{multirow}
\usepackage{subcaption} 
\usepackage{hyperref}
\hypersetup{
    colorlinks=true,
    linkcolor=blue,
    filecolor=magenta,      
    urlcolor=cyan,
}
\usepackage{xcolor}

\DeclareMathOperator*{\atan2}{atan2}
\newcommand{\norm}[1]{\left\lVert#1\right\rVert}
\newcommand{\abs}[1]{\left|#1\right|}
\newtheorem{prop}{Proposition}
\begin{document}
\title{Set-theoretic Localization for Mobile Robots with Infrastructure-based Sensing}

\author[1]{Xiao Li}

\author[1]{Yutong Li}

\author[1]{Nan Li}

\author[1]{Anouck Girard}

\author[1]{Ilya Kolmanovsky}

\authormark{Li \textsc{et al}}

\address[1]{\orgdiv{Department of Aerospace Engineering}, \orgname{The University of Michigan - Ann Arbor}, \orgaddress{\state{Michigan}, \country{USA}}}

\corres{Xiao Li, Aerospace, Robotics, and Controls Laboratory, the University of Michigan - Ann Arbor, MI-48109, USA. \email{hsiaoli@umich.edu}}

\fundingInfo{National Science Foundation under Award ECCS-1931738.}

\abstract[Summary]{In this paper, we propose a set-membership based localization approach for mobile robots using infrastructure-based sensing. Under an assumption of known uncertainties bounds of the noise in the sensor measurement and robot motion models, the proposed method computes uncertainty sets that over-bound the robot 2D body and orientation via set-valued motion propagation and subsequent measurement update from infrastructure-based sensing. We establish theoretical properties and computational approaches for this set-theoretic localization method and illustrate its application to an automated valet parking example in simulations, and to omnidirectional robot localization problems in real-world experiments. With deteriorating uncertainties in system parameters and initialization parameters, we conduct sensitivity analysis and demonstrate that the proposed method, in comparison to the FastSLAM, has a milder performance degradation, thus is more robust against the changes in the parameters. Meanwhile, the proposed method can provide estimates with smaller standard deviation values.}


\maketitle
\section{Introduction}\label{sec:intro}
\begin{figure}[t]
    \centering
    \includegraphics[width=0.9\textwidth]{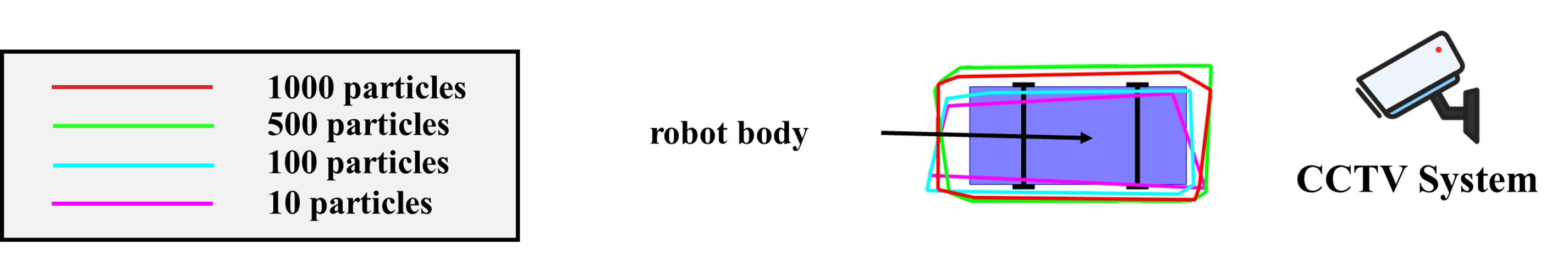}
    \caption{FastSLAM estimated robot body area.}
    \label{fig:collisionExample}
\end{figure}

One of the major challenges for navigating mobile robots safely is in their accurate and reliable localization~\cite{thrun2002probabilistic}. A promising approach is to leverage the infrastructure-based sensing and wireless communications/V2X~\cite{boban2018connected}. With the increasing computational capability of hardware, real-time simultaneous localization and mapping (SLAM) has been more widely adopted for mobile robot localization tasks in unmapped environments~\cite{thrun2007simultaneous}. In particular, with prior knowledge of the surroundings, the infrastructure-based SLAM is an appealing centralized localization approach as it reduces computational burden by treating individual agent's localization tasks independently~\cite{teixeira2010tasking}. However, quantification of the localization uncertainties is generally handled by the estimation of confidence intervals or ellipsoids within probability-based methods, e.g. Bayesian filters, particle filters~\cite{thrun2002probabilistic}, etc. Adequate explicit uncertainty bounds, which are crucial for the operation of safety-critical systems, as illustrated by the following mobile robot localization example in Fig.~\ref{fig:collisionExample}, could be difficult to generate via probabilistic methods.  

Specifically, suppose a centralized closed-circuit television (CCTV) system that collects measurements associated with the robot is set up as shown in Fig.~\ref{fig:collisionExample}, and suppose the FastSLAM~\cite{fastslam} based on particle filtering is used to estimate the area covered by the robot's body. As shown in Fig.~\ref{fig:collisionExample}, localization results with fewer particles tend to underestimate the area. If we further use the estimated area in a planning module~\cite{ceccarelli2004set}, the one which fails to contain the entire robot body may eventually cause a collision. Though the estimated areas with larger numbers of particles, e.g. 500 and 1000, over-bound the robot body, the increased sampling and computation burden impedes the online deployment of the FastSLAM. In fact, the estimated area is guaranteed to contain the robot body if and only if we sample an infinite number of particles. We also note that statistical properties of the noise and uncertainty on which probabilistic estimates depend are often assumed; however, they may not necessarily hold in practice.

In this paper, in order to obtain  quantitative deterministic uncertainty bounds on estimated states unavailable with probabilistic approaches, we extend the set-theoretic localization approach~\cite{di2004set} to the infrastructure based sensing setting~\cite{song2013cooperative,sung2011vehicle,khalid2020smart}. The proposed method can be readily adapted to both camera and lidar sensor systems. With the assumption of bounded robot dynamics uncertainty and measurement noise, the proposed approach guarantees that the actual states are necessarily within the estimated uncertainty sets, which provides a desired quantitative uncertainty bound on the estimated states. Polytopes are used to approximate the uncertainty sets to reduce conservativeness. We use the automated valet parking as an example to validate the effectiveness of the proposed method in simulations, and compare the results with the ones using the FastSLAM. Moreover, we also demonstrate that the proposed algorithm can be readily applied to real-world systems via indoor hardware experiments on a mobile robot. 

The main contributions of this paper are as follows: (1) We extend the existing set-theoretic localization approach~\cite{di2004set} to an infrastructure-based sensing setting. (2) We use polytopes to approximate the uncertainty sets which reduces conservativeness as compared to boxes~\cite{di2004set} and is still computationally efficient due to low dimensional characteristics of the problem. (3) We demonstrate the proposed algorithm in a simulated auto-valet parking and a real-world ominidirectional robot localization applications. (4) In a sensitivity analysis, compared to the FastSLAM, we demonstrate the proposed method is more robust and can provide estimates with smaller standard deviation values in presence of changes in the system parameters and initial conditions. 
\section{Related Work}\label{sec:relatedWork}
Estimation problems~\cite{barfoot2017state} associated with localization in robotics have been extensively studied. Algorithms such as the classical Kalman filter, particle filter, Bayesian filter, and unscented Kalman filter have been considered for the robotics localization problems~\cite{thrun2002probabilistic,barfoot2017state,smith2013sequential}. A comprehensive review of invariant Kalman filtering that uses the geometric structure of the state space and the dynamics to improve the performance of the extended Kalman filter (EKF) is available in the literature~\cite{barrau2018invariant}. Our method has a similar structure as the convectional filtering algorithms that predict the states via a dynamics model and subsequently update the prediction using sensor measurements. Unlike the probabilistic methods such as the EKF, the proposed algorithm uses set-valued motion prediction and measurement update to yield a deterministic estimation of uncertainty bounds for robotics localization, mapping, and system state estimation problems~\cite{hanebeck1996set,alamo2005guaranteed,merhy2020guaranteed,di2004set, di2001setThesis,wangSet,kolmanovsky2006simultaneous}.   

In scenarios of real-time exploration tasks, SLAM algorithms are necessary as the environment information is unknown to the robots~\cite{thrun2002probabilistic}. For SLAM problems, probabilistic methods, for example, EKF SLAM  \cite{moutarlier1990experimental,moutarilier1989stochastic} and FastSLAM~\cite{fastslam} have been developed and widely adopted. With the advances in computing power, matrix and graph optimization algorithms~\cite{grisetti2010tutorial}, such as iSAM~\cite{kaess2008isam} and GTSAM~\cite{gtsam}, have become feasible for real-time implementation. SLAM algorithms that rely on visual sensors e.g. monocular, stereo, or RGB-D cameras~\cite{orbslam,lsdslam,monoslam,rgbdslam} have also been proposed. In contrast to the classical SLAM problem, our method exploits an infrastructure-based sensing setting, and a prior knowledge of the environment. Moreover, our method is based on centralized localization and hence is able to reduce computational burden by treating individual agents' localization tasks independently~\cite{teixeira2010tasking}. 

A wide range of sensors and their combinations have been considered for localization applications. Localization using point clouds generated from lidar sensors has been explored~\cite{wolcott2015fast, yoneda2014lidar} and a more general review of the lidar point cloud registration algorithms was presented by Pomerleau et al.~\cite{pomerleau2015review}. As a more economical sensor option, visual cameras have been used in visual localization and visual odometry generation~\cite{nister2004visual}. With the development of machine learning, significant progress in outdoor visual place recognition and visual localization has been achieved~\cite{VPR,naseer2017semantics, pirker2011cd, NetVLAD, nister2004visual}. Meanwhile, sensor fusion such as visual-inertia odometry~\cite{leutenegger2015keyframe} and visual-lidar fusion~\cite{debeunne2020review} have proved to be effective in mobile robot localization. In our problem setting, as long as the aforementioned sensor systems can provide range and/or angle measurements, they can be directly used as the infrastructure-based sensors in our localization algorithm. 
\section{Problem Formulation And Preliminaries}\label{sec:problem}
\begin{figure}[!h]
    \centering
    \includegraphics[width=1\textwidth]{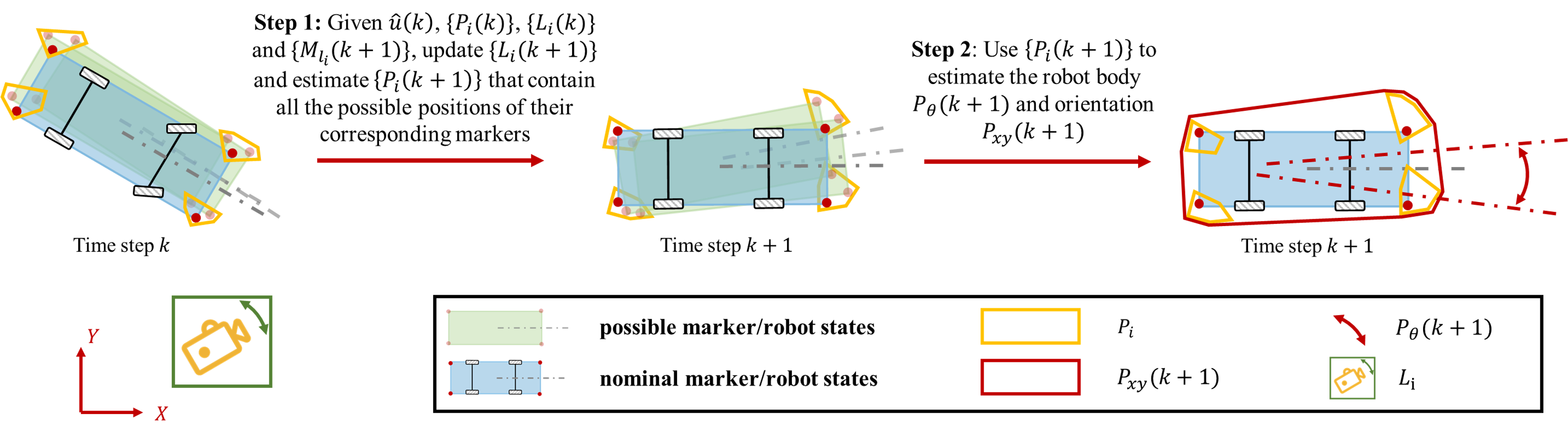}
    \caption{Illustrations of the set-theoretic localization approach.}
    \label{fig:problemFormulation}
\end{figure}

As shown in Fig.~\ref{fig:problemFormulation}, we consider a localization system in an $X-Y$ plane $\Omega\in\mathbb{R}^2$ that comprises $m$ ($m\geq 2$) infrastructure-installed sensors, where each individual sensor is located at a certain point in the $X-Y$ plane, and a mobile robot, to which $n$ identical sensor detection markers are attached such that the entire robot body is in the convex hull of the markers. We assume that the sensor system is able to measure the relative angle and/or the relative range of the markers to the sensors. If the sensors' installation positions and orientations were perfectly known, the sensor system could function as a global positioning system for the markers, and subsequently, could localize the robot body area as the convex envelope formed by the markers. However, the actual sensor position and orientation is estimated during the initial calibration~\cite{devarajan2008calibrating} and these estimates may have errors. Thus, the actual sensor's position and orientation may not be accurately known; in this paper, we assume that the $i^{th}$ sensor's actual position and orientation states are a priori known to belong to an uncertainty set $L_i\in\mathbb{R}^2\times[-\pi,\pi]$ as shown in Fig.~\ref{fig:problemFormulation}. 

We denote the actual robot body and orientation as $\hat{P}_{xy}\subset\Omega$ and $\hat{p}_{\theta}\in[-\pi,\pi]$, respectively, and represent the $i^{th}$ marker's actual position by $\hat{p}_i=[\hat{p}_{i,x}\;\hat{p}_{i,y}]^T\in\Omega$. Given robot dynamics, it is possible to derive the equations of motion for the markers. We assume that the marker dynamics can be represented by the following expressions, 
\begin{equation}\label{eq:f}
    \hat{p}_i(k+1) = f_i\left(\hat{p}_i(k), \hat{u}(k)+w_{u}(k)\right) + w_f^i(k),\; i=1,\dots,n, 
\end{equation}
where the control input $\hat{u}(k)\in\mathbb{R}^{|u|}$ is subject to an unknown additive noise $w_{u}(k)\in\mathbb{R}^{|u|}$, and $w_f^i(k)\in\mathbb{R}^{2}$ represents the unmodeled disturbance. We assume that the noise $w_{u}(k)$ and disturbance $w_f^i(k)$ are bounded such that $\abs{w_{u}(k)} \leq \epsilon^u$, $\norm{w_f^i(k)}_{\infty} \leq \epsilon^f$ with known upper bounds $\epsilon^u\in\mathbb{R}^{|u|}$ and $\epsilon^f\in\mathbb{R}$. These bounds are characterized from measurements collected during preliminary experimentation with the robots.

We denote the $i^{th}$ sensor's actual state as $\hat{l}_i=[\hat{l}_{i,x}\;\hat{l}_{i,y}\;\hat{l}_{i,\theta}]^T\in L_i$ where the actual installation position is $\hat{l}_{i,xy}=[\hat{l}_{i,x}\;\hat{l}_{i,y}]^T\in\Omega$ and the actual orientation is $\hat{l}_{i,\theta}\in[-\pi,\pi]$. Each single measurement vector $h_{i,j^*}(k) \in \mathbb{R}^{|g|}$ from the $i^{th}$ sensor obeys the following sensor model,
\begin{equation}\label{eq:y}
    h_{i,j^*}(k) = g_i\left(\hat{l}_i(k), \hat{p}_{j^*}(k) \right) + w_g^{i,j^*}(k), 
\end{equation}
where the marker identity $j^*\in \{1,\dots,n\}$ is latent as all markers are identical to the sensor system and $w_g^{i,j^*}(k)\in\mathbb{R}^{\abs{g}}$ is the unknown additive noise. We assume the noise $w_g^{i,j^*}(k)$ is bounded by a known upper bound $\epsilon^g\in\mathbb{R}^{\abs{g}}$ such that $\abs{w_{g}^{i,j^*}(k)} \leq \epsilon^g$. This bound is determined, for instance, from sensor accuracy specification by the sensor manufacturer. 

Considering all the aforementioned uncertainties in both the robot dynamics and the sensor measurements, we aim to develop an algorithm that estimates the actual robot body $\hat{P}_{xy}$ and orientation $\hat{p}_{\theta}$ based on the models~\eqref{eq:f},~\eqref{eq:y} under the above bounded noise/uncertainty assumptions. Specifically, at time step $k$, we treat the actual robot body $\hat{P}_{xy}(k)$, the actual robot orientation $\hat{p}_{\theta}(k)$, the markers' actual position  $\{\hat{p}_i(k)\}_{i=1,\dots,n}$, and the sensors' actual states $\{\hat{l}_i(k)\}_{i=1,\dots,m}$ as unknown and consider the uncertainty sets $\{P_i(k)\}_{i=1,\dots,n}$ and $\{L_i(k)\}_{i=1,\dots,m}$, where $\hat{p}_i(k)\in P_i(k)$ and $\hat{l}_i(k)\in L_i(k)$. Note that sensor uncertainties sets $L_i(0)$ can be generated through the initial calibration~\cite{devarajan2008calibrating} and the marker uncertainty sets $P_i(0)$ can be initialized by coarse localization using robot on-board sensors~\cite{bostanci2019lidar}. At time step $k+1$, the markers' positions are updated by a given control signal $\hat{u}_k$ through the dynamics in~\eqref{eq:f}. Afterwards, based on~\eqref{eq:y}, each individual $i^{th}$ sensor in the localization system produces a set of measurements $M_{l_i}(k+1)=\{h_{i,j^*}(k+1)\},\;i=1,\dots,m$. 

As shown in Fig~\ref{fig:problemFormulation}, the goals of our set-theoretic localization method are as follows:
\begin{enumerate}
    \item Given control $\hat{u}(k)$, the sensor uncertainty sets $\{L_i(k)\}_{i=1,\dots,m}$, the marker uncertainty sets $\{P_i(k)\}_{i=1,\dots,n}$, and the measurements $\{M_{l_i}(k+1)\}_{i=1,\dots,m}$, compute the sensor uncertainty sets $\{L_i(k+1)\}_{i=1,\dots,m}$ and estimate marker uncertainty sets $\{P_i(k+1)\}_{i=1,\dots,n}$ based on (\ref{eq:f}) and (\ref{eq:y}) such that $\hat{l}_{i}\in L_i(k+1)$ for $i=1,\dots,m$ and $\hat{p}_{i}(k+1)\in P_i(k+1)$ for $i=1,\dots,n$.
    \item Based on the uncertainty sets $\{L_i(k+1)\}_{i=1,\dots,m}$, $\{P_i(k+1)\}_{i=1,\dots,n}$, estimate two sets $P_{xy}(k+1)\subset\Omega$ and $P_{\theta}(k+1)\subset[-\pi,\pi]$ such that the robot body is entirely contained in the estimated set, i.e., $\hat{P}_{xy}(k+1)\subset P_{xy}(k+1)$ and the orientation is within the estimated interval, i.e., $\hat{p}_{\theta}(k+1)\in P_{\theta}(k+1)$.
\end{enumerate}
\section{Mathematical Model}\label{sec:model}
\begin{figure}[!h]
    \centering
    \includegraphics[width=0.95\textwidth]{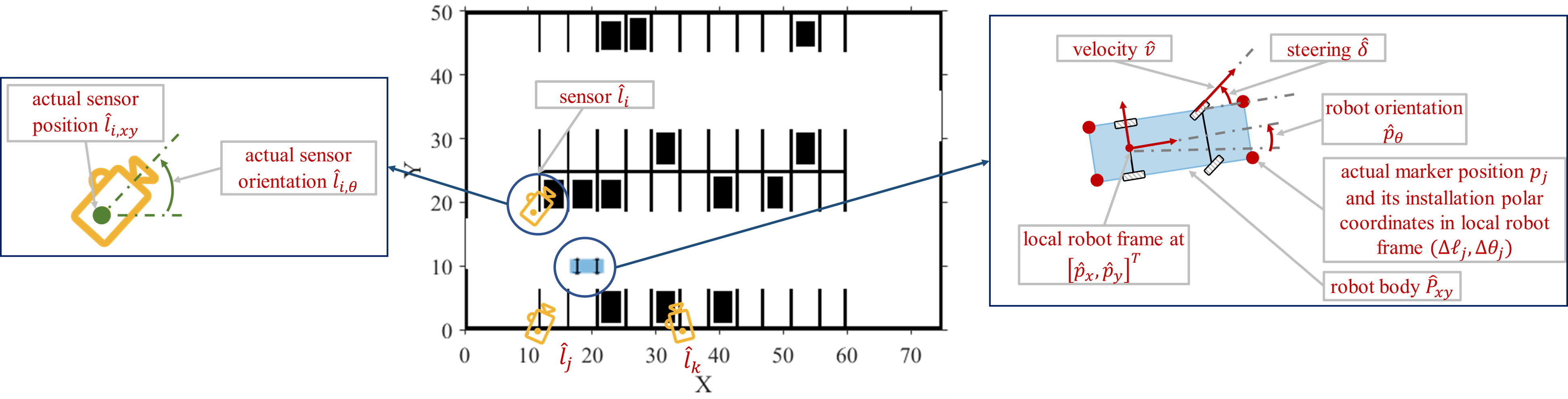}
    \caption{Modeling of an infrastructure-based localization system.}
    \label{fig:parkingSpace}
\end{figure}
In this paper, as shown in Fig.~\ref{fig:parkingSpace}, we assume the robot is a front-wheel drive vehicle that is subject to longitudinal velocity and steering control $\hat{u} = [\hat{v}\;\hat{\delta}]^T\in\mathbb{R}^2$, where the unknown additive noise $w_{u}=[w_v\; w_{\delta}]^T$ is bounded by $\epsilon^u=[\epsilon^v\; \epsilon^{\delta}]^T$, i.e., $\abs{w_{v}} \leq \epsilon^v$ and $\abs{w_{\delta}} \leq \epsilon^{\delta}$. Then, the robot kinematics can be represented by a discrete-time model,
\begin{equation}\label{eq:bicycle}
\left\{
\begin{aligned}
    \hat{p}_{\theta}(k+1) &=  \hat{p}_{\theta}(k) + \frac{(\hat{v}+w_{v})\cdot dt}{\ell}\cdot \sin{(\hat{\delta}+w_{\delta})}\\
    \hat{p}_x(k+1) &=  \hat{p}_x(k) + (\hat{v}+w_{v})\cdot dt \cdot \cos{\hat{p}_{\theta}(k)}\cdot \cos{(\hat{\delta}+w_{\delta})}  \\
    \hat{p}_y(k+1) &=  \hat{p}_y(k) + (\hat{v}+w_{v})\cdot dt \cdot \sin{\hat{p}_{\theta}(k)}\cdot \cos{(\hat{\delta}+w_{\delta})}
\end{aligned}\right.,
\end{equation}
where $dt$ is the sampling period, $\ell$ is the length of the robot wheelbase, and $[\hat{p}_x\;\hat{p}_y]^T$ is the center of the robot rear wheel axis. In what follows, we derive the markers' equations of motion~\eqref{eq:f} in Sec.~\ref{subsec:robDynamicsModel} and the sensor measurement model~\eqref{eq:y} , which generates angle and range measurements in Sec.~\ref{subsec:sensorModel}.
\subsection{Robot and Marker Kinematics Models}\label{subsec:robDynamicsModel}
Derived from \eqref{eq:bicycle}, and as a realization of \eqref{eq:f}, the kinematics of the $i^{th}$ marker are represented by
\begin{equation}\label{eq:motionModel}
    \hat{p}_i(k+1) 
    =
    \hat{p}_i(k) 
    +
    \left[
    \begin{array}{c}
        d_i(\hat{v}+w_{v}, \hat{\delta}+w_{\delta})\cdot\cos{(\theta_i(\hat{v}+w_{v}, \hat{\delta}+w_{\delta}, \hat{p}_{\theta}(k)))}\\
        d_i(\hat{v}+w_{v}, \hat{\delta}+w_{\delta})\cdot\sin{(\theta_i(\hat{v}+w_{v}, \hat{\delta}+w_{\delta}, \hat{p}_{\theta}(k)))}\
    \end{array}
    \right] + w_f^i(k),
\end{equation}
where 
\begin{equation*}
\begin{aligned}
    d_i(v,\delta) &= v\cdot dt\cdot \sqrt{(\frac{\Delta \ell_i\sin{\delta}}{\ell})^2+(\cos{\delta})^2-\frac{\Delta \ell_i}{\ell}\cdot\sin{\Delta \theta_i}\cdot\sin{(2\delta)}},\\
    \theta_i(v,\delta, \hat{p}_{\theta}(k)) &= \hat{p}_{\theta}(k) + \Delta \theta_i + \atan2\left(\Delta\ell_i\tan{\delta}-\ell\sin{\Delta \theta_i}, \ell\cos{\Delta \theta_i}\right),
\end{aligned}
\end{equation*}
$\Delta \ell_i = \sqrt{(\hat{p}_{i,x}-\hat{p}_x)^2+(\hat{p}_{i,y}-\hat{p}_y)^2}$, $\Delta \theta_i = \atan2\left(\hat{p}_{i,y}-\hat{p}_y, \hat{p}_{i,x}-\hat{p}_x\right)$, which are assumed to be given, are the polar coordinates in a local robot frame as shown in Fig.~\ref{fig:parkingSpace}. The detailed derivation is available in Appendix~\ref{sec:appendix-kinematics}.  

\subsection{Sensor Measurement Model}\label{subsec:sensorModel}
We assume that a sensor, e.g., stereo camera or lidar, is capable of producing angle measurement $\alpha$ and range measurement $r$. Consequently, each individual measurement is a vector $h_{i,j}=[\alpha_{i,j}\; r_{i,j}]^T$, which contains the angle and range measurements corresponding to an marker of unknown identity $j\in\{1,\dots,n\}$, from the $i^{th}$ sensor. As a realization of \eqref{eq:y}, the measurement model can be represented by
\begin{equation}\label{eq:measurementModel}
\left[
\begin{array}{c}
     \alpha_{i,j}(k)  \\
     r_{i,j}(k) 
\end{array}
\right]
=
\left[
\begin{array}{c}
    \atan2(\hat{p}_{j,y}(k)-\hat{l}_{i,y}(k),\;\hat{p}_{j,x}(k)-\hat{l}_{i,x}(k)) - \hat{l}_{i,\theta}(k)  \\
    \sqrt{(\hat{p}_{j,y}(k)-\hat{l}_{i,y}(k))^2+(\hat{p}_{j,x}(k)-\hat{l}_{i,x}(k))^2} 
\end{array}
\right]
+ 
\left[
\begin{array}{c}
    w_{a}^{i,j}(k)  \\
    w_{r}^{i,j}(k) 
\end{array}
\right],
\end{equation}
where the unknown additive noise $w_g^{i,j}(k)=[w_{a}^{i,j}(k)\; w_{r}^{i,j}(k)]^T\in\mathbb{R}^2$ is bounded by $\epsilon^g=[\epsilon^{w_a}\;\epsilon^{w_r}]^T\in\mathbb{R}^2$, i.e., $\abs{w_{a}^{i,j}(k)} \leq \epsilon^{w_a}$ and $\abs{w_r^{i,j}(k)} \leq \epsilon^{w_r}$. Note that in our problem setting, as long as the infrastructure-based sensor can measure the range and/or angle of markers, there is no restriction on the sensor's type, i.e., it can be a monocular camera, lidar, et al. We also note that the marker can be virtual, such as ORB~\cite{orb} or SIFT~\cite{sift} features with object detection~\cite{papageorgiou2000trainable} such as vehicle wheel detection~\cite{achler2004camera,achler2004vehicle} to refine the region of interest.
\section{Set-theoretic Localization}\label{sec:method}
For simplification of the presentation, in Sec.~\ref{subsec:propagation} and Sec.~\ref{subsec:measurementUpdate}, we assume the sensor system measures only the relative angles from the markers to the sensors, e.g. using monocular cameras, and the measurement is labeled with the corresponding marker identity. We first use the marker kinematics model to propagate the uncertainty sets (Sec.~\ref{subsec:propagation}), then update the sets with corresponding measurements derived from the infrastructure-based sensors (Sec.~\ref{subsec:measurementUpdate}). By incorporating the geometrical constraints between individual markers, we can improve the estimation accuracy of the robot body and orientation (Sec.~\ref{subsec:rigidBody+setRecon}). Then, we extend our method to the scenario where the actual measurement-to-marker correspondence is latent (Sec.~\ref{subsec:correspondance}). An extension of the proposed method to the sensor case with both range and angle measurements, e.g., using stereo cameras, is introduced in Sec.~\ref{subsec:extendStereo}. We conclude the section with a set over-approximation strategy to simplify the set operations (Sec.~\ref{subsec:polytopeSetEst}).  

\subsection{Motion Propagation}\label{subsec:propagation}
For the sensor uncertainty set propagation, we decompose $L_i$ into two bounded sets $L_{i,xy}\subset \Omega$ and $L_{i,\theta}\subset [-\pi,\pi]$ such that $L_i \subset L_{i,xy}\times L_{i,\theta}$ and $\times$ stands for the Cartesian product, which simplifies the update computations in Sec.~\ref{subsec:measurementUpdate}. Then, based on (\ref{eq:motionModel}), the uncertainty sets $P_i$ of the $i^{th}$ marker and $L_{i,xy}$, $L_{i,\theta}$ of the $i^{th}$ sensor are updated as
\begin{equation}\label{eq:markerSetUpdate}
    P_i(k+1|k) = P_i(k) \oplus \left(D_{i,x}(k)\times D_{i,y}(k)\right) \oplus \mathcal{B}_{\infty}(\epsilon^{f}),
\end{equation} 
\begin{equation}\label{eq:camSetUpdateDecomposed}
\left\{
    \begin{aligned}
    L_{i,xy}(k+1|k) &= L_{i,xy}(k)\\
    L_{i,\theta}(k+1|k) &= L_{i,\theta}(k) 
    \end{aligned}\right.,
\end{equation}
where $\oplus$ denotes the Minkowski sum, $\mathcal{B}_{\infty}(\epsilon^{f})$ is an $\infty\text{-norm}$ ball of radius $\epsilon^{f}$, and 
\begin{equation*}
\begin{aligned}
    D_{i,x}(k)&= \left\{dx\in\mathbb{R}: dx=d_i(v,\delta)\cdot\cos{(\theta_i(v,\delta,p_{\theta}))},\;\abs{v-\hat{v}}\leq\epsilon^{v},\;\abs{\delta-\hat{\delta}}\leq\epsilon^{\delta},\;p_{\theta}\in P_{\theta}(k)\right\},\\
    D_{i,y}(k)&= \left\{dy\in\mathbb{R}: dy=d_i(v,\delta)\cdot\sin{(\theta_i(v,\delta,p_{\theta}))},\;\abs{v-\hat{v}}\leq\epsilon^{v},\;\abs{\delta-\hat{\delta}}\leq\epsilon^{\delta},\;p_{\theta}\in P_{\theta}(k)\right\},
\end{aligned}
\end{equation*}
are the intervals that bound the displacements of the $i^{th}$ marker along $X$ and $Y$ axis, respectively. We note that $D_{i,x}$, $D_{i,y}$ can be computed using interval arithmetic. Equation~\eqref{eq:camSetUpdateDecomposed} reflects the fact that the sensors are stationary.

\subsection{Measurement Update}\label{subsec:measurementUpdate}
Given measurements $M_{l_i}(k+1)=\{\alpha_{i,j}\}$, $i=1,\dots, m$, we first update the uncertainty set $L_{i,\theta}$, and then we sequentially update $L_{i,xy}$ and $P_{j}$. Based on~\eqref{eq:measurementModel}, we update the sensor orientation uncertainty set as follows,
\begin{equation}\label{eq:updateCamTheta_KnownMatching}
    L_{i,\theta}(k+1) =
    L_{i,\theta}(k+1|k) \bigcap \left(\bigcap\limits_{\alpha_{i,j}\in M_{l_i}}[\psi_{i,j}, \phi_{i,j}]\right),
\end{equation}
where $\psi_{i,j}$, $\phi_{i,j}$ are derived from each individual measurement $\alpha_{i,j}$ as
\begin{equation*}\label{eq:nominalCamThetaRange}
\begin{aligned}
\psi_{i,j} &= \psi(L_{i,xy}, P_{j}, \alpha_{i,j}) = \beta_{min} - \alpha_{i,j} - \epsilon^{w_a},\\
\phi_{i,j} &= \phi(L_{i,xy}, P_{j}, \alpha_{i,j}) = \beta_{max} - \alpha_{i,j} + \epsilon^{w_a},
\end{aligned}
\end{equation*}
and $\beta_{min}$, $\beta_{max}$ are defined as follows
\begin{equation}\label{eq:betaMinMax}
\begin{aligned}
    \beta_{min} &= \min\limits_{
    \substack{l_{i,xy} \in L_{i,xy}(k+1|k),\\ 
    p_j \in P_{j}(k+1|k)}}
    \left( \atan2(p_{j,y}-l_{i,y},\,p_{j,x}-l_{i,x}) \right),\\
    \beta_{max} &= \max\limits_{
    \substack{l_{i,xy} \in L_{i,xy}(k+1|k),\\ 
    p_j \in P_{j}(k+1|k)}}
    \left( \atan2(p_{j,y}-l_{i,y},\,p_{j,x}-l_{i,x}) \right)
\end{aligned}
\end{equation}
with bounded sets $L_{i,xy}(k+1|k)$ and $P_{j}(k+1|k)$.

Subsequently, with the updated $L_{i,\theta}(k+1)$ and measurements $M_{l_i}(k+1)$, $i=1,\dots,m$, we estimate the uncertainty sets $L_{i,xy}$ and $P_j$ as follows
\begin{equation}\label{eq:updateCamXY_KnownMatching}
    L_{i,xy}(k+1) = L_{i,xy}(k+1|k) \bigcap 
    \left(\bigcap\limits_{\alpha_{i,j}\in M_{l_i}}\left( P_{j}(k+1|k) \oplus L_{M}(p_j, \alpha_{i,j})\right)\right),
\end{equation}
\begin{equation}\label{eq:updateMarkerXY_KnownMatching}
    P_{j}(k+1) = P_{j}(k+1|k) \bigcap
    \left(\bigcap\limits_{\substack{i=1,\dots,m,\\ \alpha_{i,j}\in M_{l_i}}} \left(\left( L_{i,xy}(k+1) \oplus  P_{M}(l_i, \alpha_{i,j})\right)\right)\right),
\end{equation}
where the sets $L_{M}$ and $P_{M}$ are defined as
\begin{equation}\label{eq:measurementSetLocalAtMarker}
   L_{M}(p_j, \alpha_{i,j}) :=   \{[l_x\;l_y]^T\in\mathbb{R}^2|\abs{\alpha_{i,j}-\atan2(-l_{y}, -l_{x})+\theta_c}\leq \epsilon^{w_a} + \delta \theta_c\},  
\end{equation}
\begin{equation}\label{eq:measurementSetLocalAtCam}
    P_{M}(l_i, \alpha_{i,j}) := 
    \{[p_x\;p_y]^T\in\mathbb{R}^2 |\abs{\alpha_{i,j}-\atan2(p_{y}, p_{x})+\theta_c}\leq \epsilon^{w_a} + \delta \theta_c\},
\end{equation}
with $\theta_c-\delta \theta_c$, $\theta_c+\delta \theta_c$ being the minimum and maximum of $L_{i,\theta}(k+1)$, respectively. In fact, in the coordinate frame with $p_j$ as the origin, the set $L_{M}(p_j, \alpha_{i,j})$ represents a feasible region of $l_i$ for which measurement $\alpha_{i,j}$ is plausible. Analogously, in a reference frame centered at the $i^{th}$ camera, the $j^{th}$ marker should belong to the set $P_{M}(l_i, \alpha_{i,j})$ given the measurement $\alpha_{i,j}$.

\subsection{Robot Body and Orientation Estimation}\label{subsec:rigidBody+setRecon}
Before estimating the robot body and orientation, we can exploit a rigid body constraint for two arbitrary markers $\hat{p}_i,\hat{p}_j\in\{\hat{p}_i\}_{i=1,\dots,n}$ of the form $\norm{\hat{p}_i-\hat{p}_j}_2=r_{ij}$ to further reduce the sizes of the uncertainty sets as follows
\begin{equation}\label{eq:rigidBodyRefine}
    \begin{aligned}
        P_i(k+1) &= P_i(k+1)\cap\left(P_j\oplus\mathcal{B}_2(r_{ij})\right),\\
        P_j(k+1) &= P_j(k+1)\cap\left(P_i\oplus\mathcal{B}_2(r_{ij})\right),
    \end{aligned}
\end{equation}
where $\mathcal{B}_2(r_{ij})$ represents a $\ell_2\text{-norm}$ ball of radius $r_{ij}$. 

\begin{figure}[!ht]
\centering
\includegraphics[width=0.65\linewidth]{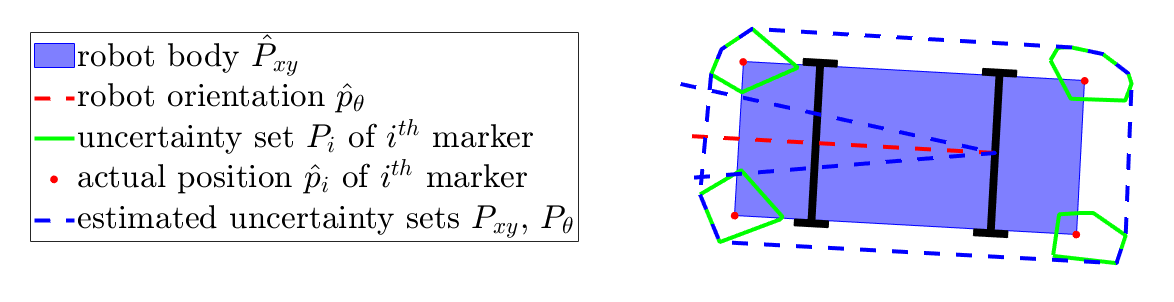}
\caption{Robot body estimation using convex envelope of marker uncertainty sets.}
\label{fig:convhullRecon}
\end{figure}

Based on the set propagation in (\ref{eq:markerSetUpdate}), (\ref{eq:camSetUpdateDecomposed}), measurement updates in (\ref{eq:updateCamTheta}), (\ref{eq:updateCamXY}), (\ref{eq:updateMarkerXY}) and set refinement by rigid body constraint in \eqref{eq:rigidBodyRefine}, the robot body is over-bounded by a convex envelope $P_{xy}(k+1)$ as shown in Fig.~\ref{fig:convhullRecon} which can be represented as
\begin{equation}\label{eq:positionSetRecon}
    P_{xy}(k+1):={\tt convHull}(\{P_i(k+1)\}_{i=1,\dots,n}),
\end{equation}
where {\tt convHull()} is a standard set operation that computes a convex envelope of the given sets. The convex hull computations can be realized using set computational toolboxes, e.g. CORA~\cite{cora1, cora2, cora3}. Consequently, we can update the robot orientation uncertainty set as follows
\begin{equation}\label{eq:headingSetRecon}
    P_{\theta} = \bigcap\limits_{i,j=1,\dots,n,\;i\neq j}[\underline{\beta}_{ij}-\Delta\theta_{ij},\overline{\beta}_{ij}-\Delta\theta_{ij}]
\end{equation}
where $\Delta\theta_{ij}$ is the offset angle between the vector from $\hat{p}_i$ to $\hat{p}_j$ and the actual robot orientation $\hat{p}_{\theta}$, and where
\begin{equation*}
\begin{aligned}
    \underline{\beta}_{ij} &= \min\limits_{
    \substack{l_{i,xy} \in L_{i,xy}(k+1),\\ 
    p_j \in P_{j}(k+1)}}
    \left( \atan2(p_{j,y}-l_{i,y},\,p_{j,x}-l_{i,x}) \right),\\
    \overline{\beta}_{ij} &= \max\limits_{
    \substack{l_{i,xy} \in L_{i,xy}(k+1),\\ 
    p_j \in P_{j}(k+1)}}
    \left( \atan2(p_{j,y}-l_{i,y},\,p_{j,x}-l_{i,x}) \right).
\end{aligned}
\end{equation*}
\subsection{Latent Measurement-to-marker Correspondence}\label{subsec:correspondance}
With the assumption of latent measurement-to-marker correspondence, we first deduce the possible correspondence solutions, then, the proposed method presented in Sec.~\ref{subsec:propagation}-\ref{subsec:rigidBody+setRecon} is changed to use modified set updates in \eqref{eq:updateCamTheta_KnownMatching}, \eqref{eq:updateCamXY_KnownMatching} and \eqref{eq:updateMarkerXY_KnownMatching}. Consider an ordered set of measurements $M_{l_i}(k+1)=\{\alpha^{(q)}_{i,j'}(k+1)\}_{q=1,\dots,\abs{M_{l_i}}}$ at time step $k+1$ by the $i^{th}$ sensor that contains $\abs{M_{l_i}}\leq n$ numbers of measurements, the superscript $(q)$ indicates the order of each individual measurement in the measurement queue and each measurement $\alpha^{(q)}_{i,j'}$ has a unique correspondence to one of the $n$ markers. However, due to the measurement noise in \eqref{eq:measurementModel} and the state estimation uncertainties in \eqref{eq:markerSetUpdate} and \eqref{eq:camSetUpdateDecomposed}, it's possible that one measurement $\alpha^{(q)}_{i,j'}$ becomes feasible to multiple markers  (e.g., $j'\in\{j^*,j_1,j_2,\dots\}$) where the the actual measurement-to-marker correspondence (e.g., $j'=j^*$) is latent. Furthermore, we introduce a matrix $C_{l_i}$ of size $\abs{M_{l_i}}\times n$ to include all possible measurement-to-marker correspondence solutions given estimated uncertainty sets where $C_{l_i}(q,j)=1\text{ or }0$ indicates the $q^{th}$ measurement $\alpha^{(q)}_{i,j'}\in M_{l_i}$ can or cannot be a feasible measurement of the $j^{th}$ marker. For instance, suppose we have two measurements for four markers from the $i^{th}$ sensor, and we can represent all possible measurement-to-marker correspondence using a matrix
\begin{equation*}
    C_{l_i}(k+1) = \left[
    \begin{array}{ccccc}
        0 & 0 & 1 & 0\\
        1 & 0 & 1 & 1
    \end{array}\right],
\end{equation*}
which implies the first measurement $\alpha^{(1)}_{i,j_1}$ can be associated with the third marker, i.e., $j_1=3$, and the second measurement $\alpha^{(2)}_{i,j_2}$ is a possible measurement corresponding to the first, third and fourth markers, i.e., $j_2\in\{1,3,4\}$. Based on the definition of $C_{l_i}$, we can deduce possible measurement-to-marker correspondence that are self-consistent and mutually exclusive, i.e., only one entry equals to 1 in each row and column of the matrix $C_{l_i}$. Again, in the aforementioned example, there are two possible measurement-to-marker correspondences as follows
\begin{equation*}
    C_{l_i}^{(1)}(k+1) = \left[
    \begin{array}{ccccc}
        0 & 0 & 1 & 0\\
        1 & 0 & 0 & 0
    \end{array}\right],\;
    C_{l_i}^{(2)}(k+1) = \left[
    \begin{array}{ccccc}
        0 & 0 & 1 & 0\\
        0 & 0 & 0 & 1
    \end{array}\right],
\end{equation*}
and the actual correspondence must be one of these. 

To obtain $C_{l_i}(k+1)$ from $M_{l_i}(k+1)$, consider an individual measurement $\alpha^{(q)}_{i,j'}\in M_{l_i}(k+1)$, we note that any marker $p_j$ that satisfies the following condition is a candidate to measurement $\alpha^{(q)}_{i,j'}$:
\begin{equation}\label{eq:nonemptyMatchingCondition}
    P_j(k+1|k)\bigcap\left(L_{i,xy}(k+1|k) \oplus  P'_{M}(l_i, \alpha^{(q)}_{i,j'}) \right)\neq \varnothing,
\end{equation}
where
\begin{equation}\label{eq:PMSetMatching}
    P'_{M}(l_i, \alpha^{(q)}_{i,j'}) := \{[p_{x}\;p_{y}]^T\in\mathbb{R}^2\;|\;\abs{\alpha^{(q)}_{i,j'}-\atan2(p_{y}, p_{x})+\theta_0}\leq \epsilon^{w_a} + \delta \theta\},
\end{equation}
and $\theta_0-\delta \theta$, $\theta_0+\delta \theta$ are the minimum and maximum of $L_{i,\theta}(k+1|k)$, respectively. This is due to the fact that the marker associated with the $q^{th}$ measurement is necessarily within $L_{i,xy}(k+1|k) \oplus P'_{M}(l_i, \alpha^{(q)}_{i,j'})$. Then, we can obtain $C_{l_i}(k+1),\;i=1,\dots,m$ by examining the criteria in~\eqref{eq:nonemptyMatchingCondition} for all measurements obtained from every sensor. We can enumerate through all the possible unique solutions contained in $C_{l_i}(k+1)$ according to the principles of mutual exclusivity and logical self-consistency as in the aforementioned example. This way, for each $M_{l_i}(k+1)$, we are able to generate one or multiple measurement-to-marker correspondence solutions $C_{l_i}^{(\mu)}(k+1),\;\mu=1,\dots,c$, which contain the actual measurement-to-marker correspondence. 

Given multiple matching solutions $C^{(\mu)}_{l_i}(k+1)$, $\mu\in\{1,\dots,c\}$ for $M_{l_i}(k+1)=\{\alpha^{(q)}_{i,j'}(k+1)\}$, the update procedures for $L_{i,\theta}(k+1)$ and $L_{i,xy}(k+1)$ are similar to \eqref{eq:updateCamTheta_KnownMatching}, \eqref{eq:updateCamXY_KnownMatching} by applying union operation over all $C^{(\mu)}_{l_i}$, $\mu\in\{1,\dots,c\}$ so that $\hat{l}_{i,\theta}$, $\hat{l}_{i,xy}$ are necessarily within 
\begin{equation}\label{eq:updateCamTheta}
    L_{i,\theta}(k+1) = L_{i,\theta}(k+1|k) \bigcap \left(\bigcup\limits_{\mu=1}^{c}\left(\bigcap\limits_{\substack{q,\; \alpha^{(q)}_{i,j'}\in M_{l_i},\\C^{(\mu)}_{l_i}(q,j)=1}}[\psi(L_{i,xy}, P_{j}, \alpha^{(q)}_{i,j'}), \phi(L_{i,xy}, P_{j}, \alpha^{(q)}_{i,j'})]\right)\right),
\end{equation}
\begin{equation}\label{eq:updateCamXY}
    L_{i,xy}(k+1) = L_{i,xy}(k+1|k) \bigcap \left(  \bigcup\limits_{\mu=1}^{c}\left( \bigcap\limits_{\substack{q,\; \alpha^{(q)}_{i,j'}\in M_{l_i},\\C^{(\mu)}_{l_i}(q,j)=1}}\left(P_{j}(k+1|k) \oplus L_{M}(p_j, \alpha^{(q)}_{i,j'}) \right)\right)\right),
\end{equation}
respectively. For update of $P_{j}(k+1)$, given multiple correspondence solutions $C_{l_i}^{(\mu)}(k+1),\;\mu=1,\dots,c$ for $M_{l_i}$, we suppose there is no corresponding measurement of the $j^{th}$ marker in the actual correspondence $C^{(\mu^*)}_{l_i}$, i.e., $\not\exists q\leq n$, such that $C_{l_i}^{(\mu^*)}(q, j)=1$. In this case, all the measurements in $M_{l_i}$ are irrelevant to the $j^{th}$ marker, which is supposed to be filtered out by the algorithm. In our framework, one can only conclude $M_{l_i}$ certainly contains measurement of the $j^{th}$ marker if there is a corresponding measurement in all correspondence solutions, i.e., $M_{l_i}\in M^j$ and $M^j=\{M_{l_i}, i=1,\dots,m\;|\;\forall\mu\in\{1,\dots,c\},\; \exists q\leq n,\; C_{l_i}^{(\mu)}(q, j)=1\}$. Consider all $M_{l_i}\in M^j,\; i=1,\dots,m$, the actual marker position $\hat{p}_{j}$ is necessarily within 
\begin{equation}\label{eq:updateMarkerXY}
    P_{j}(k+1) = P_{j}(k+1|k) \bigcap \left(\bigcap\limits_{\substack{i=1,\dots,m,\\M_{l_i}\in M^j}} \left( L_{i,xy}(k+1) \oplus \bigcup\limits_{\substack{\mu=1,\dots,c,\\\alpha^{(q)}_{i,j'}\in M_{l_i},\;C^{(\mu)}_{l_i}(q,j)=1}} P_{M}(l_i, \alpha^{(q)}_{i,j'})  \right)\right).
\end{equation}

\subsection{Sensors with Range and Angle Measurements}\label{subsec:extendStereo}
With angle and range measurements, the measurement update process of $L_{i,xy}$ and $P_j$ follows Sec.~\ref{subsec:measurementUpdate} and~\ref{subsec:correspondance} where $L_M$ in  \eqref{eq:measurementSetLocalAtMarker}, $P_M$ in \eqref{eq:measurementSetLocalAtCam} and $P_M'$ in \eqref{eq:PMSetMatching} are redefined as 
\begin{equation*}
\begin{aligned}
    &L_{M}(p_j, \alpha_{i,j}, r_{i,j}) :=\\
    &\left\{[l_x\;l_y]^T\in\mathbb{R}^2\bigg|
    \begin{array}{cc}
        |\alpha_{i,j}-\atan2(-l_{y}, -l_{x})+\theta_c|&\leq \epsilon^{w_a} + \delta \theta_c\\ 
        \abs{r_{i,j} - \sqrt{l_{x}^2+l_{y}^2}}&\leq \epsilon^{w_r}
    \end{array}
    \right\},   
\end{aligned}
\end{equation*}
\begin{equation*}
\begin{aligned}
    &P_{M}'(l_i, \alpha_{i,j}, r_{i,j}) := \\
    &\left\{[p_x\;p_y]^T\in\mathbb{R}^2\bigg| 
    \begin{array}{cc}
        |\alpha_{i,j}-\atan2(p_{y}, p_{x})+\theta_c|&\leq \epsilon^{w_a} + \delta \theta_c,\\ \abs{r_{i,j} - \sqrt{p_{x}^2+p_{y}^2}}&\leq \epsilon^{w_r}
    \end{array}
    \right\},
\end{aligned}
\end{equation*}
\begin{equation*}
\begin{aligned}
    &P_{M}(l_i, \alpha_{i,j}, r_{i,j}) := \\
    &\left\{[p_x\;p_y]^T\in\mathbb{R}^2\bigg| 
    \begin{array}{cc}
        |\alpha_{i,j}-\atan2(p_{y}, p_{x})+\theta_0|&\leq \epsilon^{w_a} + \delta \theta,\\ \abs{r_{i,j} - \sqrt{p_{x}^2+p_{y}^2}}&\leq \epsilon^{w_r}
    \end{array}
    \right\}.
\end{aligned}
\end{equation*}
The proposed method in Sec.~\ref{subsec:propagation}-\ref{subsec:extendStereo} has the following property (the proof is available in Appendix~\ref{sec:appendix-proof}):

\begin{prop}\label{prop1}
Assume $\hat{l}_{i,xy}(0) \in L_{i,xy}(0),\; \hat{l}_{i,\theta}(0) \in L_{i,\theta}(0),\; \hat{p}_j(0) \in P_{j}(0)$ and $i=1,\dots,m,\; j=1,\dots,n$. Then, based on the set-theoretic method in (\ref{eq:markerSetUpdate}), (\ref{eq:camSetUpdateDecomposed}), (\ref{eq:updateCamTheta}), (\ref{eq:updateCamXY}), (\ref{eq:updateMarkerXY}), (\ref{eq:rigidBodyRefine}), (\ref{eq:positionSetRecon}) and (\ref{eq:headingSetRecon}), the actual robot body $\hat{P}_{xy}(k)$ and orientation $\hat{p}_{\theta}(k)$ are confined to the estimated uncertainty sets $P_{xy}(k),\;P_{\theta}(k)$ for all $k>0$, i.e., $\hat{P}_{xy}(k)\subset P_{xy}(k)$,  $\hat{p}_{\theta}(k)\in P_{\theta}(k),\;\forall k\geq 0$.
\end{prop}

Proposition 1 ensures that the proposed method, based on the aforementioned bounded uncertainty/disturbance assumptions, is able to estimate uncertainty sets that bound the actual robot body and orientation at all times.

\subsection{Uncertainty Set Approximation}\label{subsec:polytopeSetEst}
\begin{figure}[!ht]
    \centering
    \includegraphics[width=0.45\textwidth]{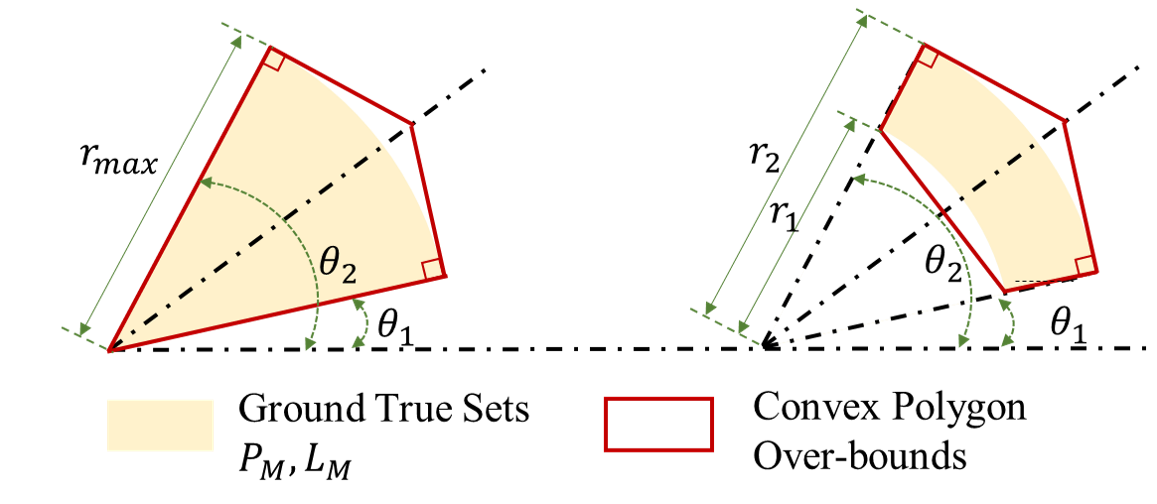}
    \caption{Illustrations of set approximations by convex polygons.}
    \label{fig:polygonEst}
\end{figure}

In this paper, we approximate the uncertainty sets using polytopes instead of boxes~\cite{di2004set}. The benefit of this approximation is that it reduce the conservativeness while this decrease of conservativeness comes at the cost of the increased computational time and effort (which is less of a concern for infrastructure-based computations in this paper as compared to onboard computations). The approximation is illustrated in Fig.~\ref{fig:polygonEst}, where $L_M$ in \eqref{eq:measurementSetLocalAtMarker} and $P_M$ in \eqref{eq:measurementSetLocalAtCam} can be represented by two circular sectors in the angular sensor system, e.g., if monocular cameras are used, and two annular sectors in the angular and range sensor system, e.g., if stereo cameras are used. We use convex polygons to over-bound $L_M$ and $P_M$ where $\theta_{1,2}=(\alpha_{i,j}+\theta_c)\pm (\epsilon^{w_a}+\delta\theta_c)$ and $r_{1,2}=r_{i,j}\pm\epsilon^{w_r}$ in Fig.~\ref{fig:polygonEst} are the angle span and range span of the (circular/annular) sectors, respectively. 

\section{Simulation and Experimental Results}\label{sec:results}
\begin{figure}[h]
    \centering
    \includegraphics[width=0.75\textwidth]{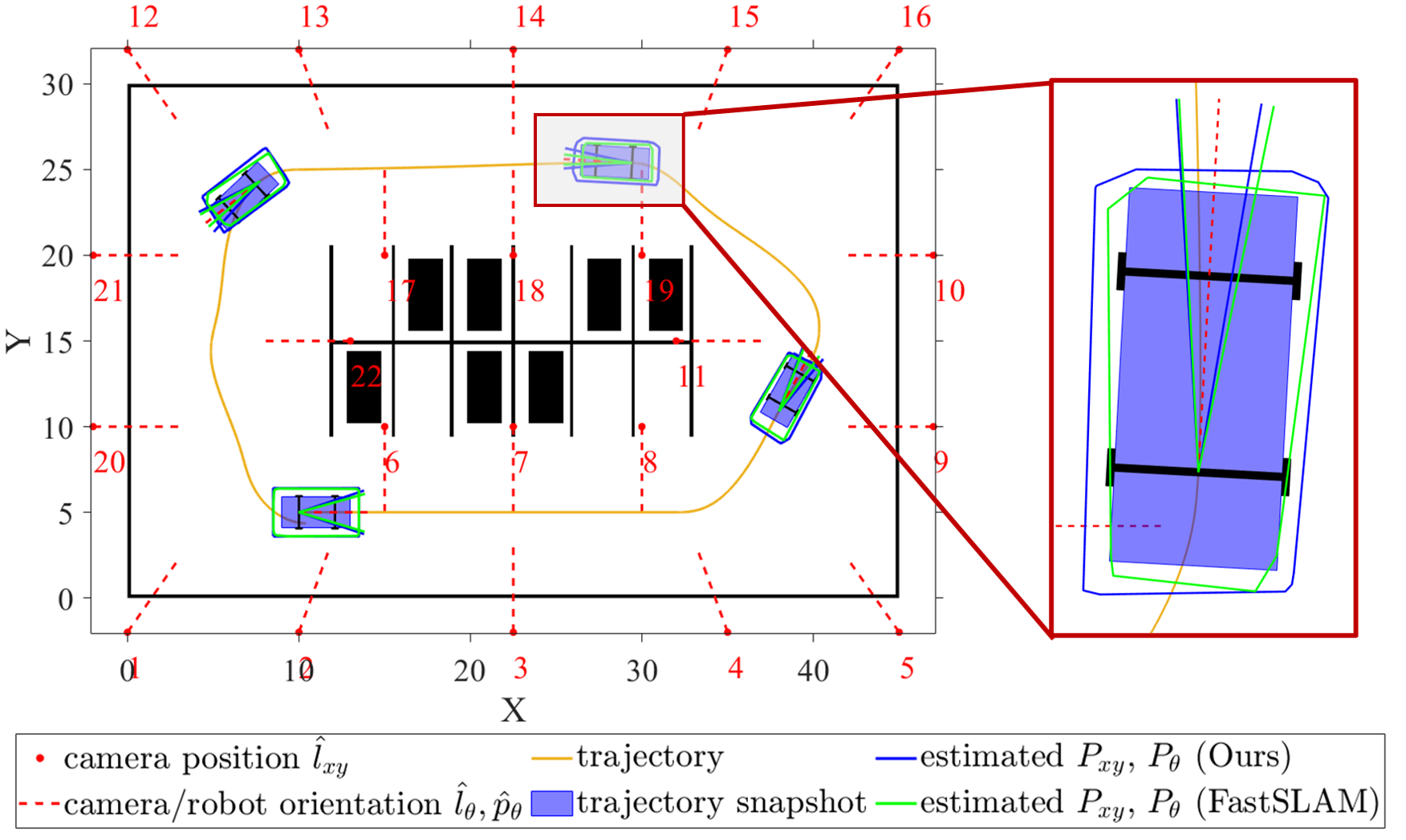}
    \caption{Simulation results of the vehicle tracking a reference trajectory in a parking space.}
    \label{fig:snapshots}
\end{figure}
In this section, we apply the proposed set-theoretic localization method to an automated valet parking example (Fig.~\ref{fig:snapshots}). As illustrated in Fig.~\ref{fig:snapshots}, a simulated parking space equipped with 21 sensors that are assumed to be stereo cameras is built based on the Automated Parking Valet toolbox~\cite{toolboxmatlab} in MATLAB. Since the vehicle's equations of motion in the simulation follow the kinematics in~\eqref{eq:bicycle}, we assume no unmodeled disturbance in~\eqref{eq:motionModel}, i.e., $w_f^i=0$. Meanwhile, each camera has a $70^{\circ}$ field of view and a $20\;\rm m$ maximum measurement range. We assume the actual measurement-to-marker correspondence is latent to the sensor system. A vehicle of length $c_l = 4\;\rm m$, width $c_w = 1.8\;\rm m$ and wheelbase $\ell=2.1\;\rm m$ is navigating within the parking space tracking a reference trajectory. We assume that four identical markers, denoted as $p_i$, $i = 1,2,3,4,$ are attached to the four vertices of the rectangle that is the vehicle's body. We apply the proposed method to localize the vehicle body $\hat{P}_{xy}$ and orientation $\hat{p}_{\theta}$, with a sampling period of $dt=0.5\;\rm s$, and the results are compared with the ones using the FastSLAM~\cite{thrun2002probabilistic}. 

The initialization and the detailed operations during the iterations of the proposed algorithm and the FastSLAM are discussed in Sec.~\ref{subsec:algorithmSetting}. We quantitatively compare the localization performances of our algorithm against the FastSLAM in Sec.~\ref{subsec:singleRun}. Then, the sensitivity analysis results of the proposed algorithm to the system and initialization parameters are discussed in Sec.~\ref{subsec:sensitivityUnknown}. A simulation example, where the proposed method is shown to mitigate the uncertainties in the sensors' orientations and positions, is presented in Sec.~\ref{subsec:sensorUpdate}. Moreover, the real-world experimental results with an omnidirectional robot and lidar-based infrastructure sensing system are presented in Sec.~\ref{subsec:lidarLoca}. The code and demonstration videos are available in \url{https://github.com/XiaoLiSean/SetThmSLAM}.

\subsection{Initialization and Iteration of Algorithms}\label{subsec:algorithmSetting}
For the initialization of the proposed method, we initialize the uncertainty sets $\{L_{i,xy}\}_{i=1,\dots,m}$, $\{P_{i}\}_{i=1,\dots,n}$ as boxes,
$\{L_{i,\theta}\}_{i=1,\dots,m}$ as intervals centered at their actual states $\{l_{i,xy}\}_{i=1,\dots,m}$, $\{p_{i}\}_{i=1,\dots,n}$, $\{l_{i,\theta}\}_{i=1,\dots,m}$, respectively. The FastSLAM is initialized with 100 particles. Each particle independently stores camera and marker states, i.e., $\{l^{(s)}_{i,xy},l^{(s)}_{i,\theta}\}_{i=1,\dots,m}$, $\{p^{(s)}_{i}\}_{i=1,\dots,n}$, $s=1,\dots,100$, which are randomly sampled from the aforementioned uncertainty sets, i.e., $\{L_{i,xy},L_{i,\theta}\}_{i=1,\dots,m}$, $\{P_{i}\}_{i=1,\dots,n}$. 

During iterations, we use CORA~\cite{cora1, cora2, cora3} in MATLAB to implement set operations between polytopes, e.g. Minkowski sums, intersections of polytopes, etc. In the FastSLAM, we estimate $P_{xy}$ using the command {\tt enclosePoints()} in CORA to compute a convex polygon that encloses all marker points $\{p_1^{(s)},p_2^{(s)},p_3^{(s)},p_4^{(s)}\}_{s=1,\dots,100}$ stored in the particles. From the snapshot in Fig. \ref{fig:snapshots}, the estimated uncertainty set using the FastSLAM (green solid line) at times fails to contain the entire vehicle body. In contrast, the proposed method guarantees that the vehicle body is always contained within the estimated set (blue solid line), which is consistent with Proposition~1. 

\subsection{Estimation Performance}\label{subsec:singleRun}
\begin{figure}[h]
\centering
\begin{tabular}[c]{ccc}
    \begin{subfigure}[c]{0.33\textwidth}
        \includegraphics[width=\linewidth]{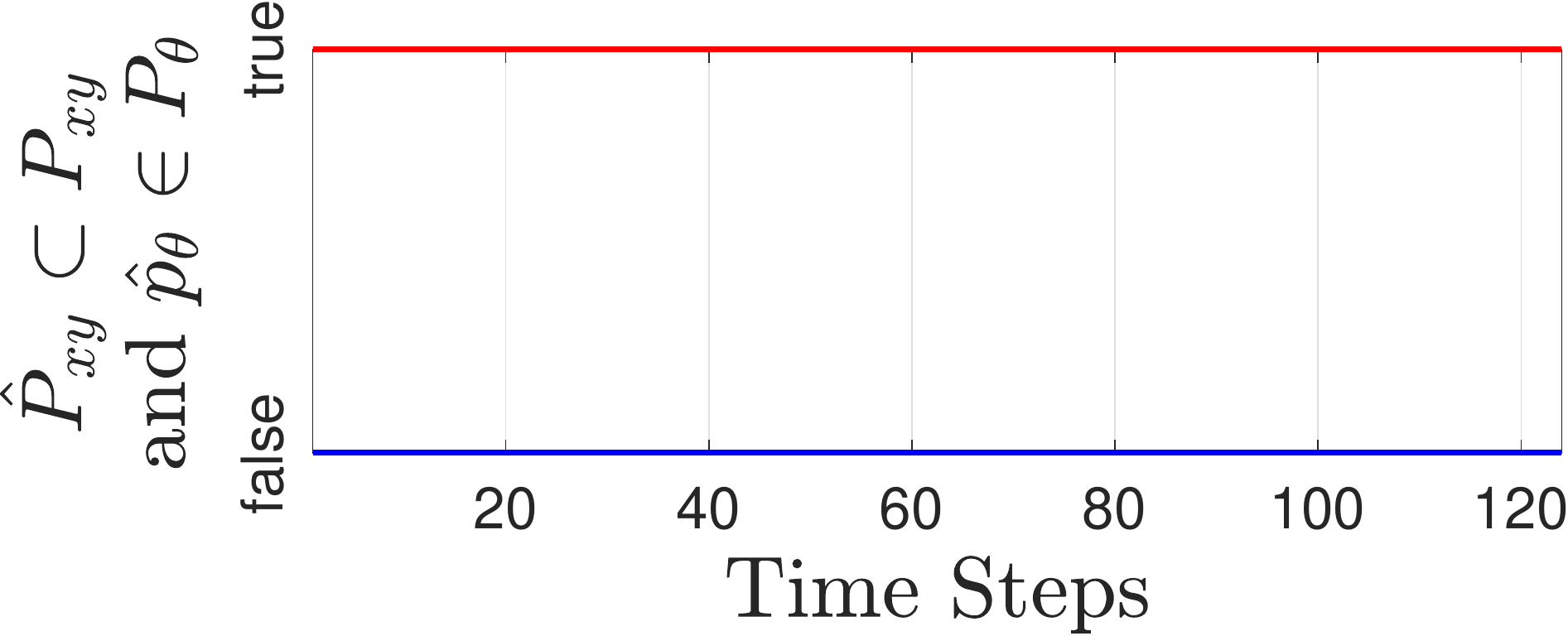}
        \caption{}
        \label{fig:isInSingleRun_Stereo}
    \end{subfigure}    
    &
    \begin{subfigure}[c]{0.33\textwidth}
        \includegraphics[width=\linewidth]{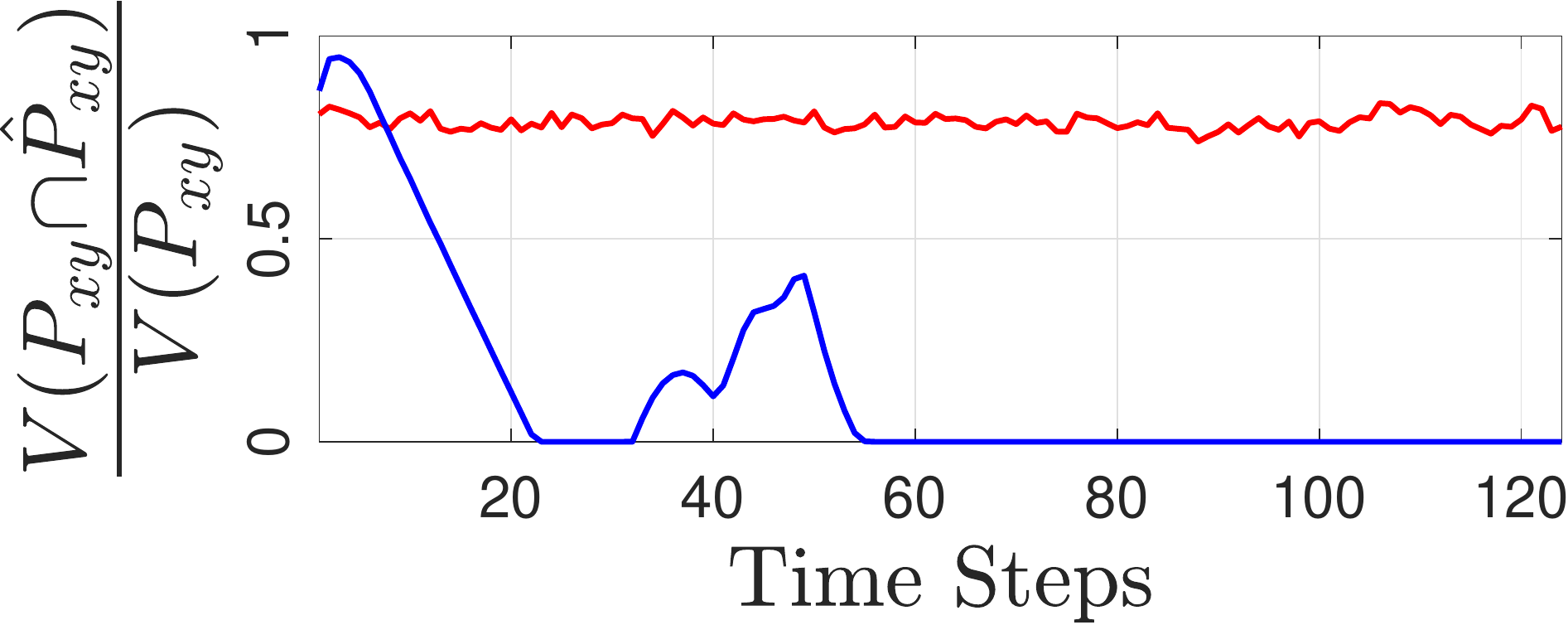}
        \caption{}
        \label{fig:pxySingleRun_Stereo_metric}
    \end{subfigure} 
    &
    \multirow{2}{*}[14pt]{
    \begin{subfigure}[c]{0.33\textwidth}
    \vspace{1cm}
        \includegraphics[width=0.82\linewidth]{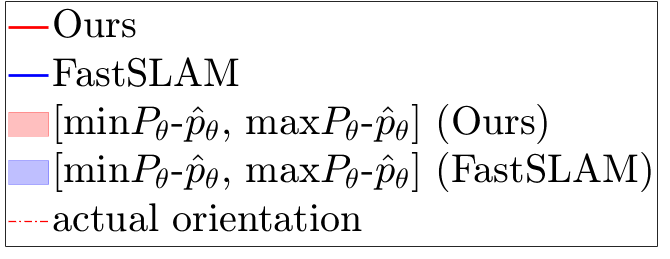}
    \end{subfigure} 
    }
    \\
    \begin{subfigure}[c]{0.33\textwidth}
        \includegraphics[width=\linewidth]{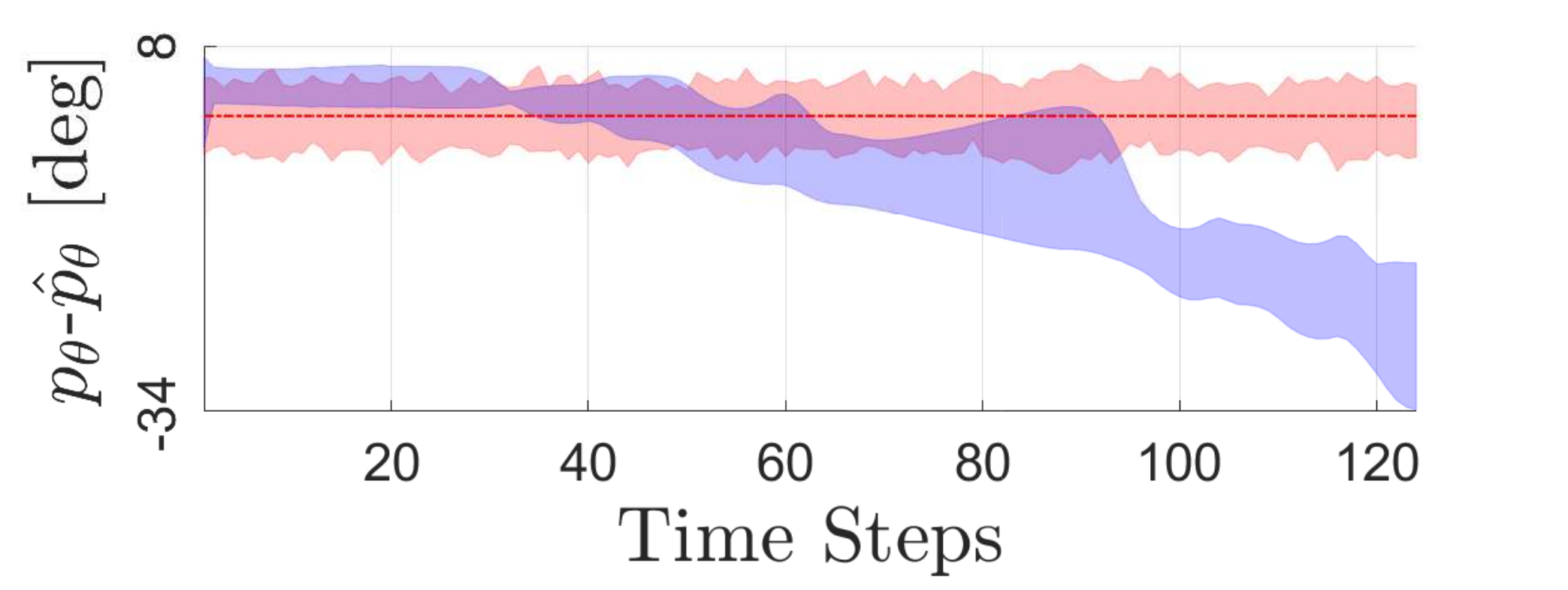}
        \caption{}
        \label{fig:ptSingleRun_Stereo}
    \end{subfigure}      
    &
    \begin{subfigure}[c]{0.35\textwidth}
        \includegraphics[width=\linewidth]{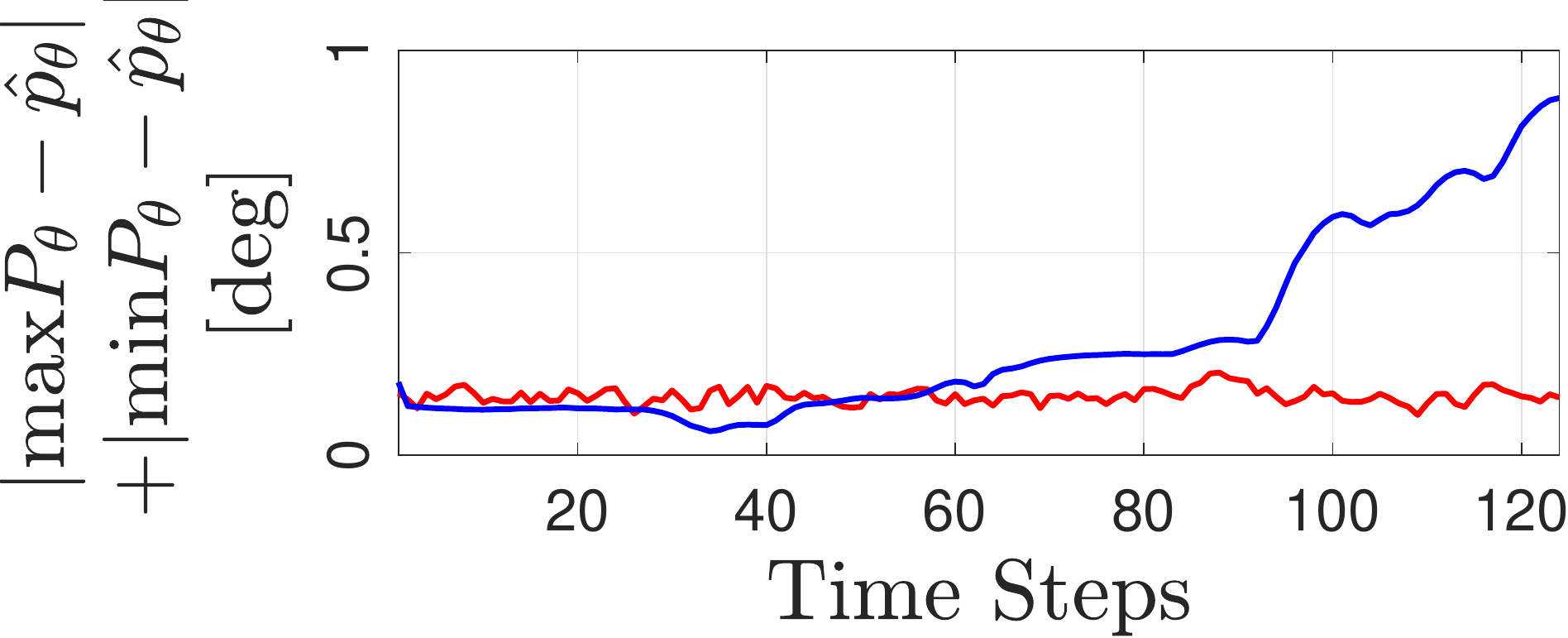}
        \caption{}
        \label{fig:ptSingleRun_Stereo_metric}
    \end{subfigure}  
    &
\end{tabular}
\caption{Comparison of state estimates between the proposed method and the FastSLAM in a single vehicle path around the parking space: (a) Logical value encoding if the actual vehicle body $\hat{P}_{xy}$ and orientation $\hat{p}_{\theta}$ are in the corresponding estimated sets. (b) Volume of intersection between estimated and actual vehicle body $V(P_{xy}\cap\hat{P}_{xy})$ divided by $V(P_{xy})$. (c) Deviation between the estimated and actual vehicle orientation. The width of the strip represents the size of the estimated uncertainty set. (d) Sum of $\abs{\max P_{\theta} -\hat{p}_{\theta}}$ (i.e., the deviation between the maximum in $P_{\theta}$ and the actual vehicle orientation) and $\abs{\min P_{\theta} -\hat{p}_{\theta}}$ (i.e., the deviation between the minimum in $P_{\theta}$ and the actual vehicle orientation).}
\label{fig:singleRun}
\end{figure}

At time step $k=0$, for the proposed method, we initialize $\{P_{i}\}_{i=1,\dots,n}$ and $\{L_{i,xy}\}_{i=1,\dots,m}$ as boxes of size $V(P_i)=1\;\rm m^2$ and $V(L_{i,xy})=0.01\;\rm m^2$, respectively, and $\{L_{i,\theta}\}_{i=1,\dots,m}$ as intervals of size $V(L_{i,\theta})=2\;\rm deg$. The particles in the FastSLAM randomly sample their states from the initialized boxes and intervals above. The markers' equations of motion are subject to noises $w_{v}$, $w_{\delta}$ with bounds  $\epsilon^{v}=0.1\;\rm m/s$ and $\epsilon^{\delta}=0.5\;\rm deg$, respectively. We assume the angle and range measurement noises of each cameras are bounded by $\epsilon^{w_a}=1\;\rm deg$, $\epsilon^{w_r}=0.1\;\rm m$. In addition, two metrics, namely $$m_1=\frac{V(P_{xy}\cap\hat{P}_{xy})}{V(P_{xy})},$$ 
$$m_2=\abs{\text{max} P_{\theta}- \hat{p}_{\theta}}+\abs{\text{min} P_{\theta}- \hat{p}_{\theta}},$$ 
are used to evaluate the estimation performance of the algorithm. For $m_1\in[0,1]$, which is used for evaluating vehicle body estimation performance, the closer the value is to 1, the better is the estimation performance. Similarly for metric $m_2\in[0, +\infty)$, which is used for evaluating vehicle orientation estimation performance, the closer the value is to 0, the better is the estimation performance.

As illustrated in Fig.~\ref{fig:isInSingleRun_Stereo}, our algorithm preserves the claimed property in Proposition 1, i.e., both actual vehicle body and orientation are guaranteed to be contained within the estimated uncertainty sets computed by the proposed method. Furthermore, as shown in Fig.~\ref{fig:pxySingleRun_Stereo_metric}, the proposed method has a more steady and higher value of $\frac{V(P_{xy}\cap\hat{P}_{xy})}{V(P_{xy})}$ compared with the one using the FastSLAM. This is attributed to the containment property of the uncertainty sets estimated by the proposed algorithm, while the uncertainty sets computed by the FastSLAM tend to drift away from the actual ones (in fact, there is a zero overlaps between the estimated and actual vehicle bodies after 60 steps). Similar results are observed in the orientation estimation, as shown in Figs.~\ref{fig:ptSingleRun_Stereo} and~\ref{fig:ptSingleRun_Stereo_metric}, the estimates generated by the proposed algorithm (red strip) contain the red dash line, which indicates $\hat{p}_{\theta}(k)\in P_{\theta}(k),\;\forall k\geq 0$. Meanwhile, the results by the FastSLAM fail to contain it and gradually deviate from the actual orientation line. Differently from the probabilistic methods, e.g., the FastSLAM, the proposed set theoretic localization method computes the uncertainty sets via deterministic set-valued motion propagation and measurement update so that no estimation biases occur, which is easily induced through the weight-based re-sampling procedure in the FastSLAM. This can be further verified by the observation that the strip by the proposed method distributes more evenly around the actual orientation line compared to the one by the FastSLAM in Fig.~\ref{fig:ptSingleRun_Stereo}. 

\subsection{Sensitivity Analysis}\label{subsec:sensitivityUnknown}
\begin{figure}[h]
\centering
\begin{tabular}[c]{ccc}
    \begin{subfigure}[c]{0.33\textwidth}
        \includegraphics[width=\linewidth]{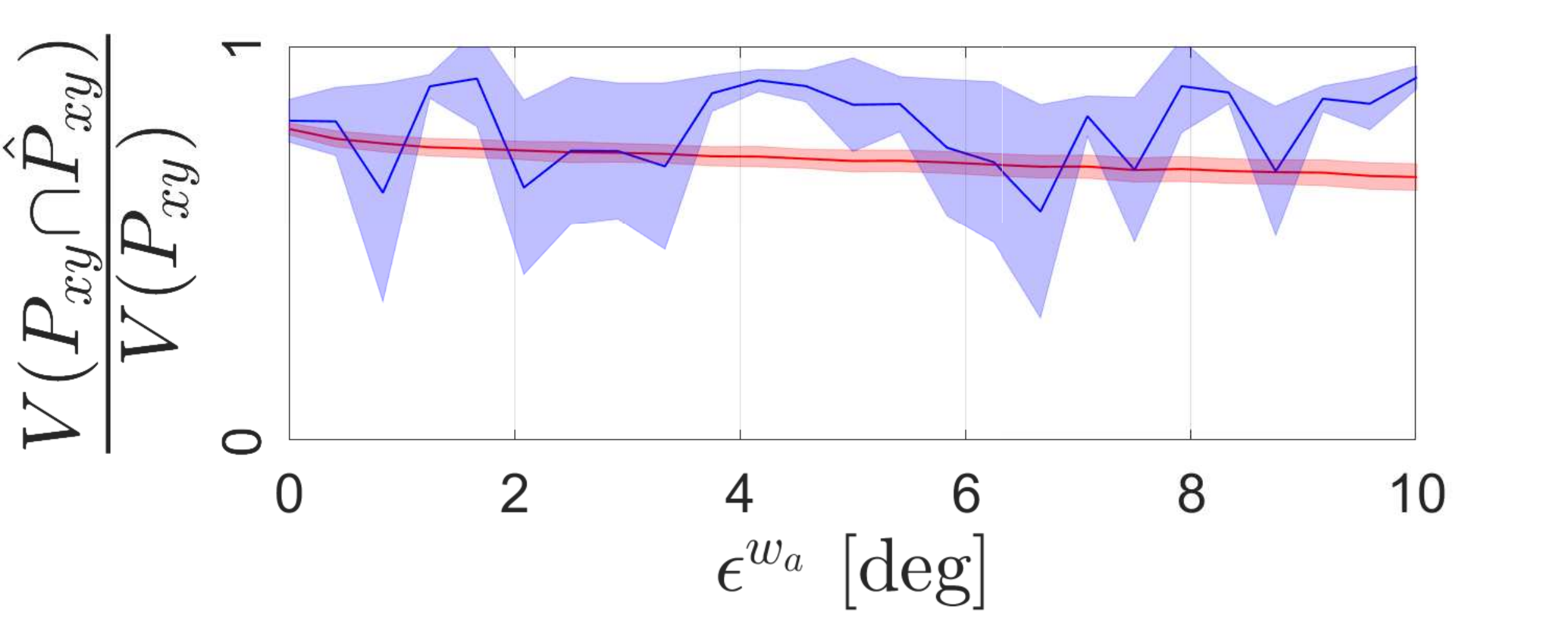}
        \caption{}
        \label{fig:sensitivityUnknown_Stereo_EWA_P}
    \end{subfigure}    
    &
    \begin{subfigure}[c]{0.33\textwidth}
        \includegraphics[width=\linewidth]{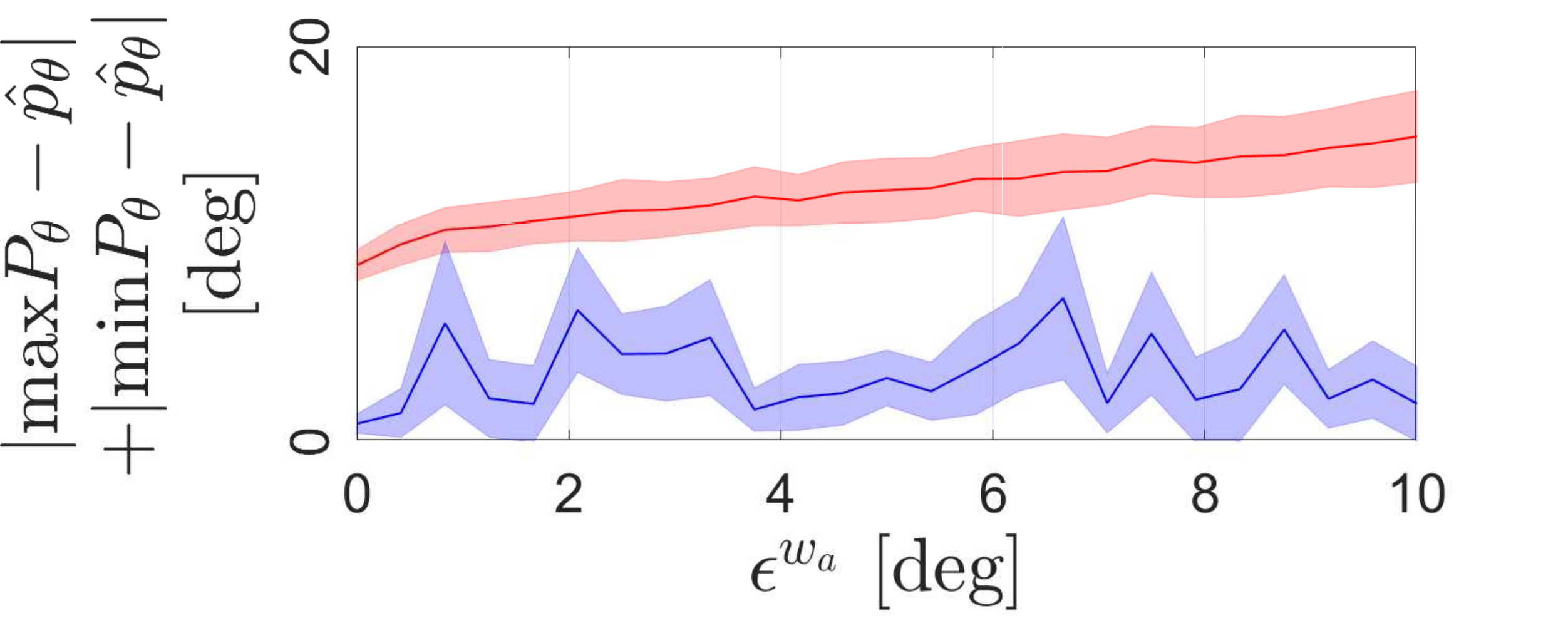}
        \caption{}
        \label{fig:sensitivityUnknown_Stereo_EWA_t}
    \end{subfigure} 
    &
    \multirow{2}{*}[14pt]{
    \begin{subfigure}[c]{0.33\textwidth}
    \vspace{1cm}
        \includegraphics[width=0.82\linewidth]{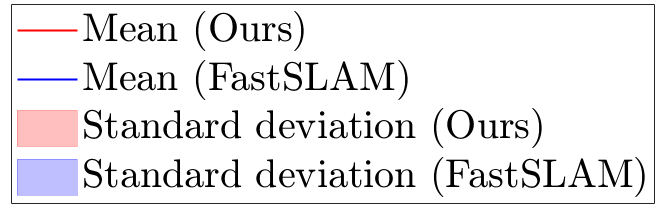}
    \end{subfigure} 
    }
    \\
    \begin{subfigure}[c]{0.33\textwidth}
        \includegraphics[width=\linewidth]{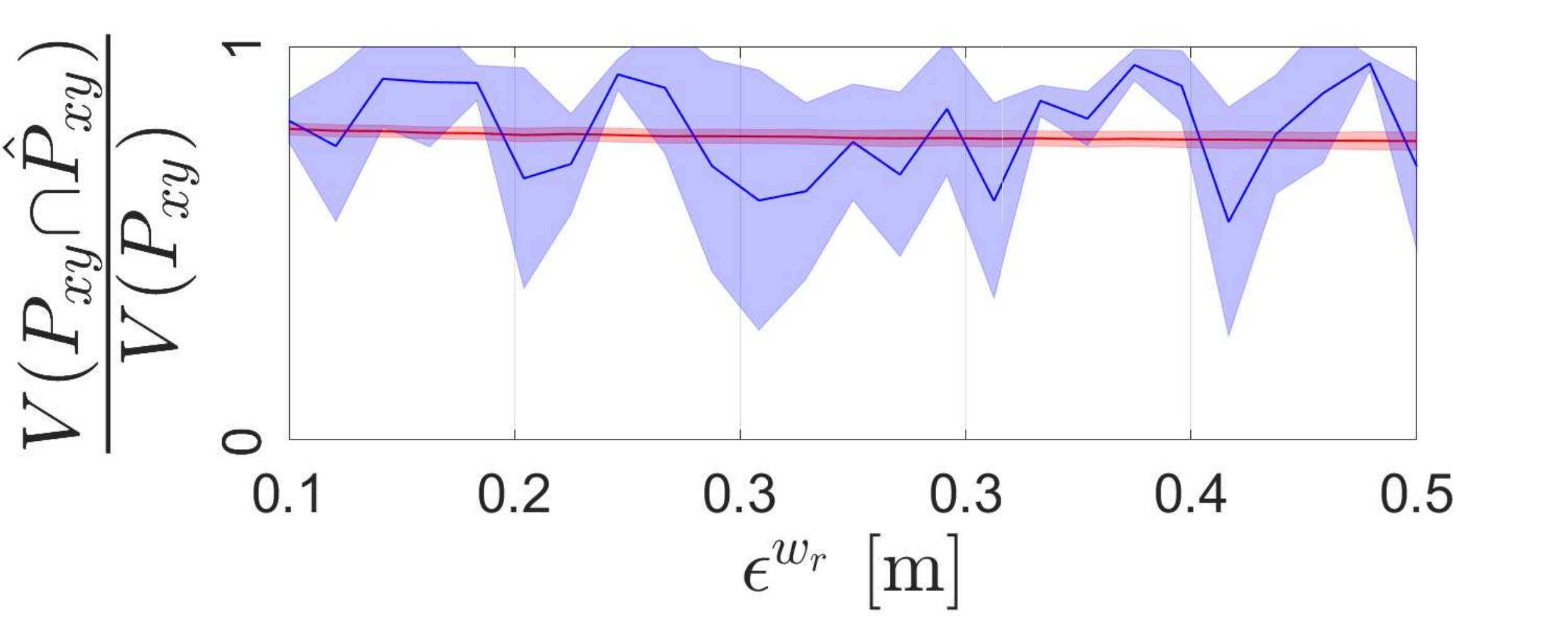}
        \caption{}
        \label{fig:sensitivityUnknown_Stereo_EWR_P}
    \end{subfigure}      
    &
    \begin{subfigure}[c]{0.35\textwidth}
        \includegraphics[width=\linewidth]{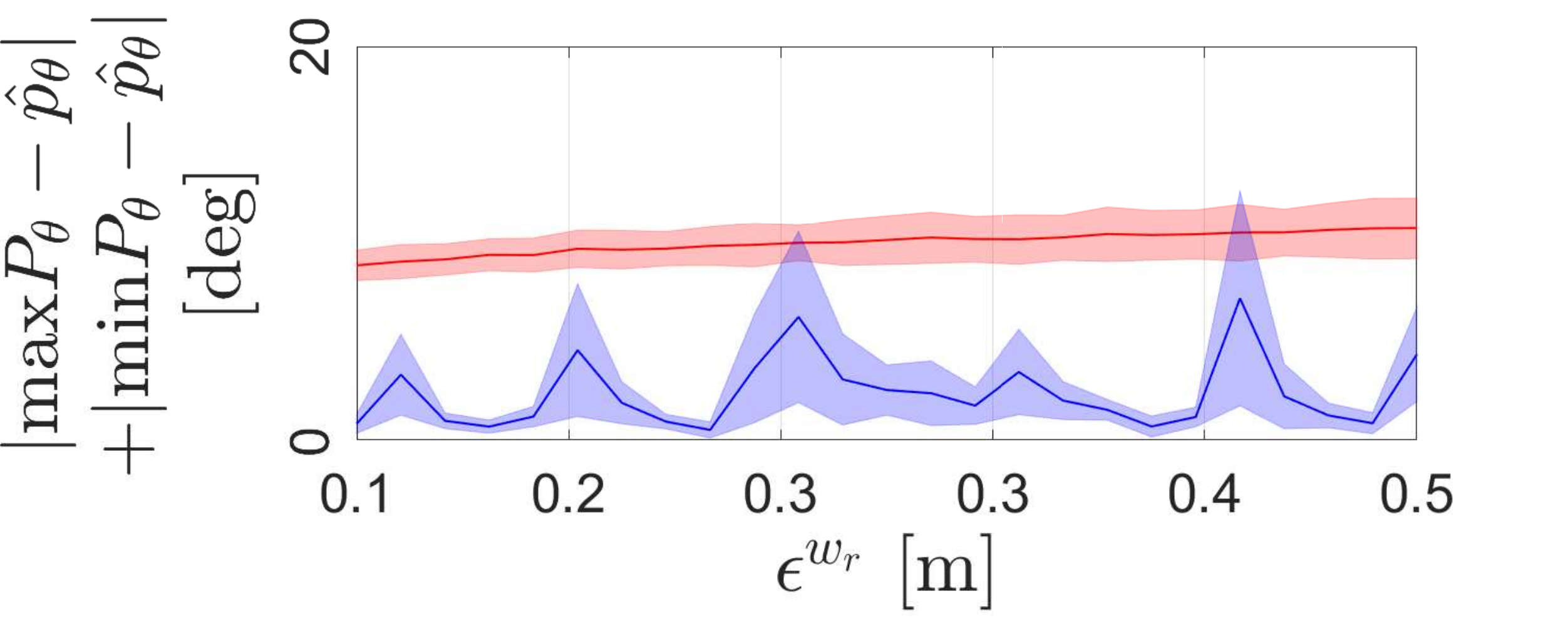}
        \caption{}
        \label{fig:sensitivityUnknown_Stereo_EWR_t}
    \end{subfigure}  
    &
\end{tabular}
\caption{Sensitivity analysis to measurement noise bounds: (a)  vehicle body estimation with different angle measurement noise bounds $\epsilon^{w_a}$. (b) vehicle orientation estimation with different angle measurement noise bounds $\epsilon^{w_a}$. (c)  vehicle body estimation with different range measurement noise bounds $\epsilon^{w_r}$. (d) vehicle orientation estimation with different range measurement noise bounds $\epsilon^{w_r}$.}
\label{fig:sensitivityMeasurementParams}
\end{figure}

\begin{figure}[h]
\centering
    \begin{subfigure}[c]{0.33\textwidth}
        \includegraphics[width=\linewidth]{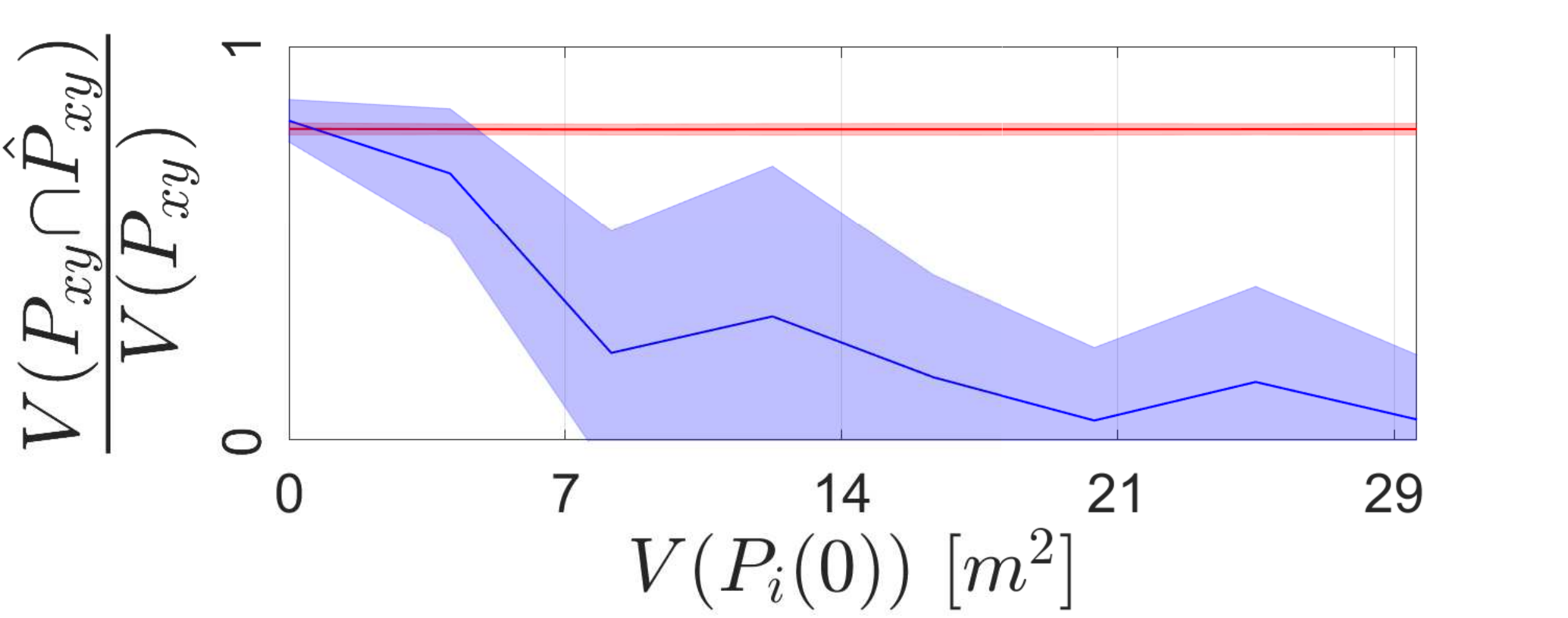}
        \caption{}
        \label{fig:sensitivityUnknown_Stereo_Pi_P}
    \end{subfigure}    
    \begin{subfigure}[c]{0.33\textwidth}
        \includegraphics[width=\linewidth]{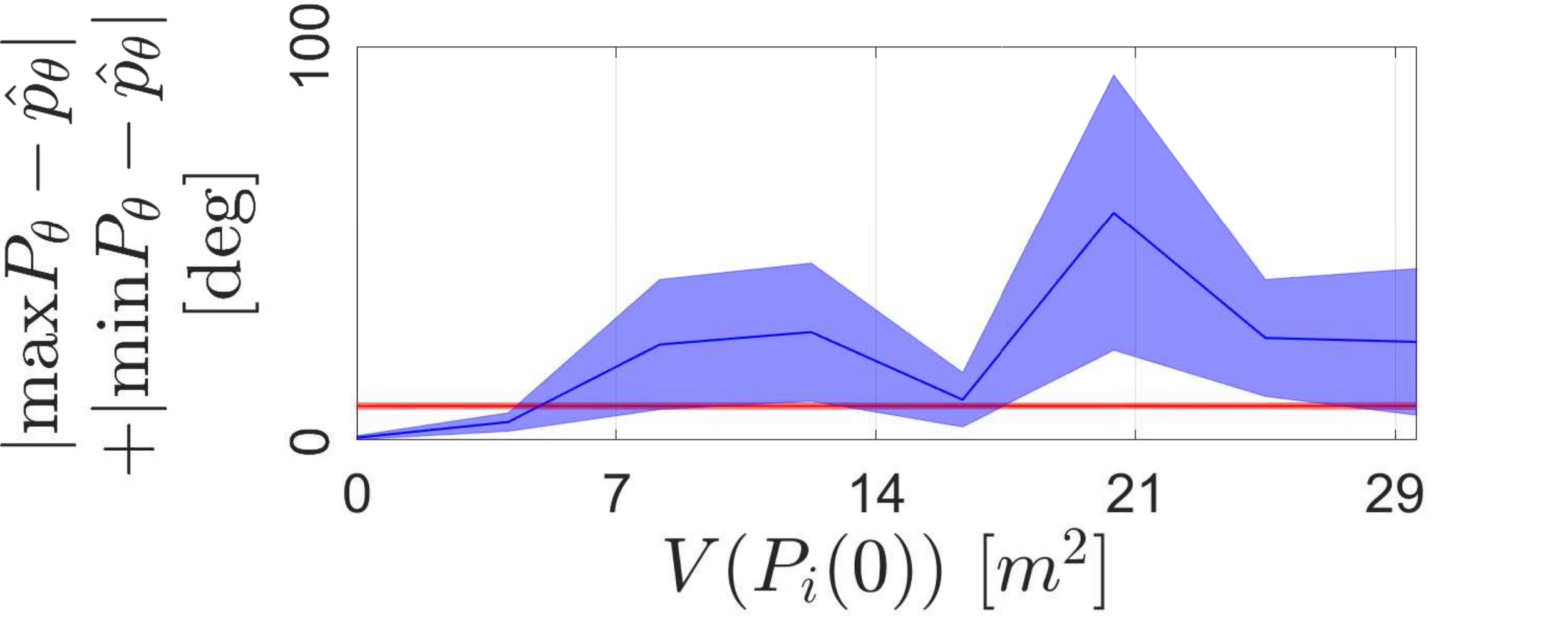}
        \caption{}
        \label{fig:sensitivityUnknown_Stereo_Pi_t}
    \end{subfigure} 
    \begin{subfigure}[c]{0.33\textwidth}
        \includegraphics[width=0.82\linewidth]{fig/legendOneCol.png}
    \end{subfigure} 
\caption{Sensitivity analysis to initial marker position uncertainties $V(P_{i}(0))$: (a) vehicle body estimation with different $V(P_{i}(0))$. (b) vehicle orientation estimation with different $V(P_{i}(0))$.}
\label{fig:sensitivityInitializationParams}
\end{figure}

\begin{figure}[h]
\centering
\begin{tabular}[c]{ccc}
    \begin{subfigure}[c]{0.33\textwidth}
        \includegraphics[width=\linewidth]{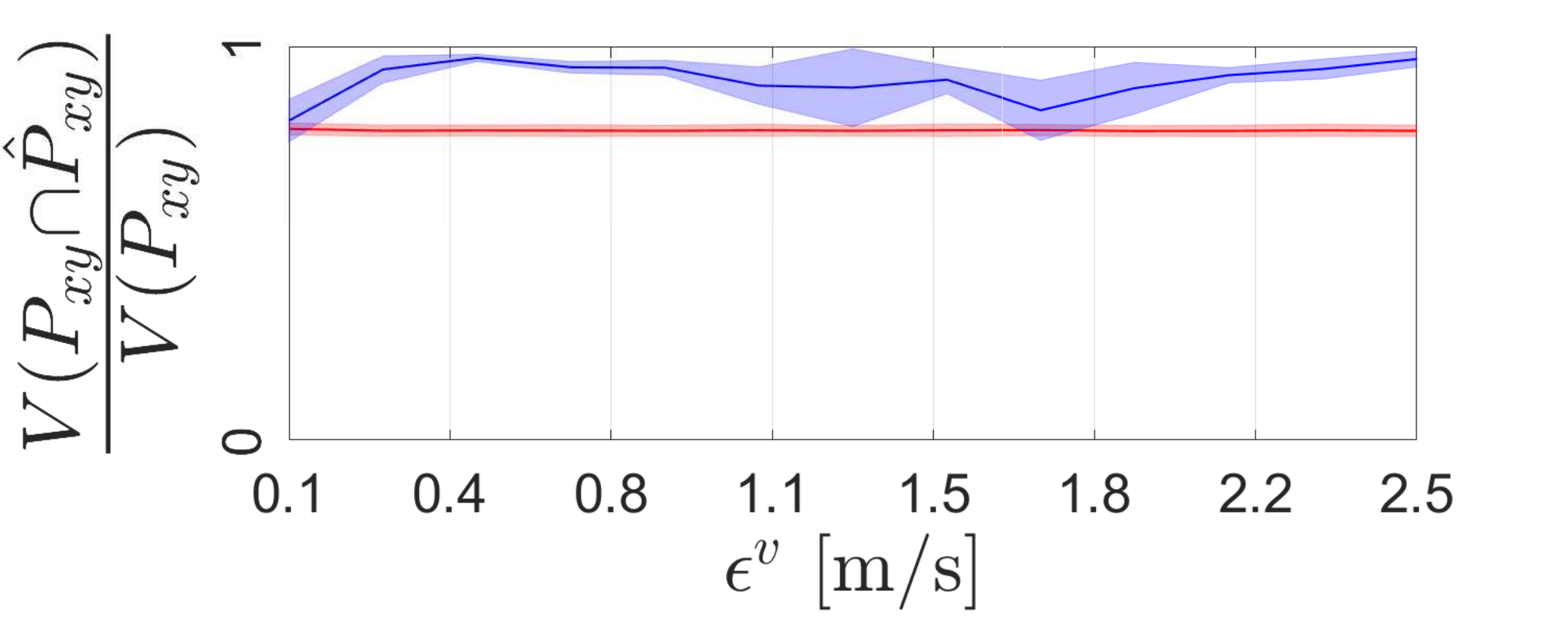}
        \caption{}
        \label{fig:sensitivityUnknown_Stereo_V_P}
    \end{subfigure}    
    &
    \begin{subfigure}[c]{0.33\textwidth}
        \includegraphics[width=\linewidth]{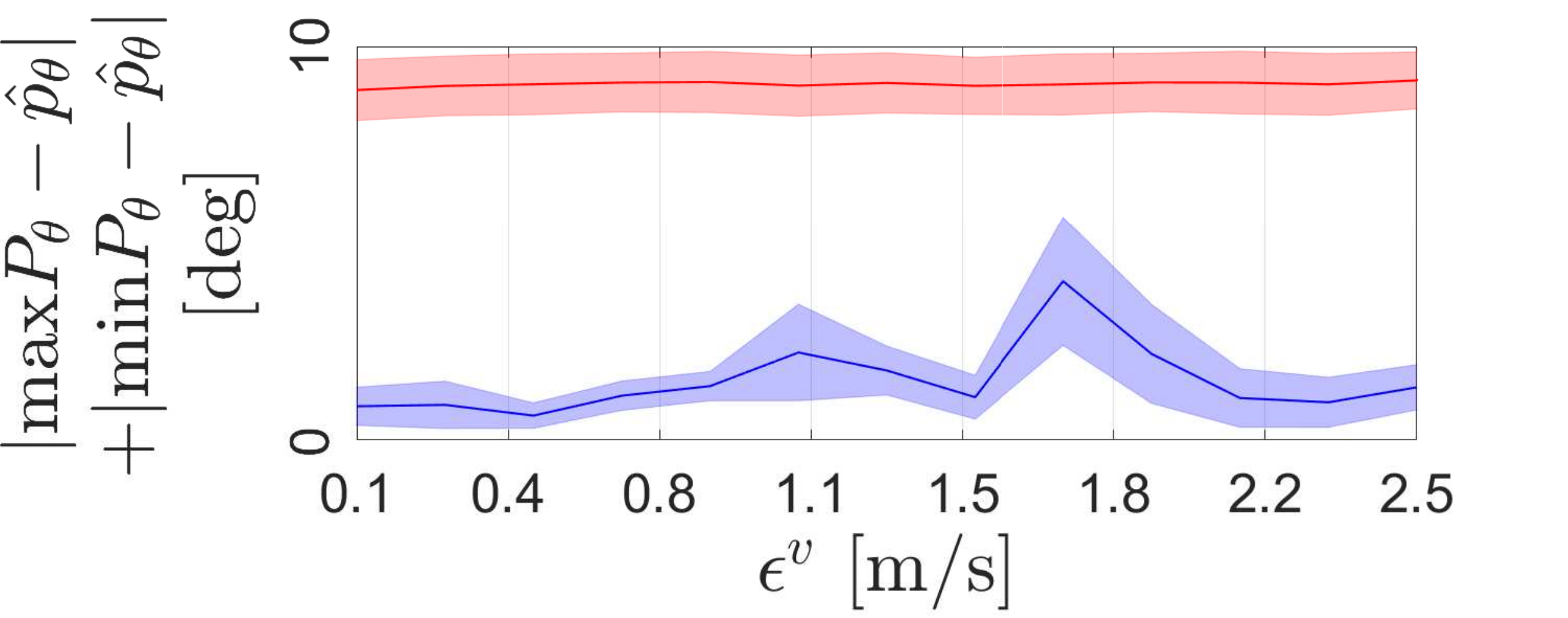}
        \caption{}
        \label{fig:sensitivityUnknown_Stereo_V_t}
    \end{subfigure} 
    &
    \multirow{2}{*}[14pt]{
    \begin{subfigure}[c]{0.33\textwidth}
    \vspace{1cm}
        \includegraphics[width=0.82\linewidth]{fig/legendOneCol.png}
    \end{subfigure} 
    }
    \\
    \begin{subfigure}[c]{0.33\textwidth}
        \includegraphics[width=\linewidth]{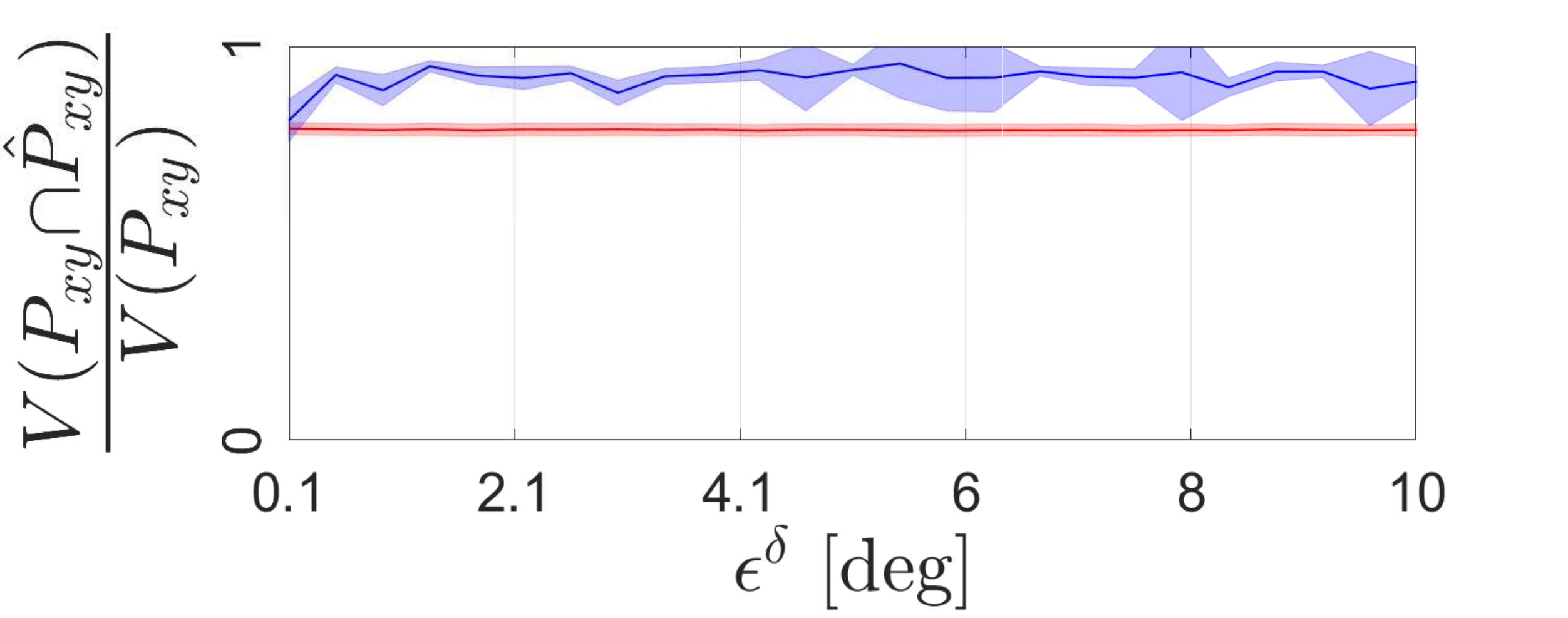}
        \caption{}
        \label{fig:sensitivityUnknown_Stereo_Delta_P}
    \end{subfigure}      
    &
    \begin{subfigure}[c]{0.35\textwidth}
        \includegraphics[width=\linewidth]{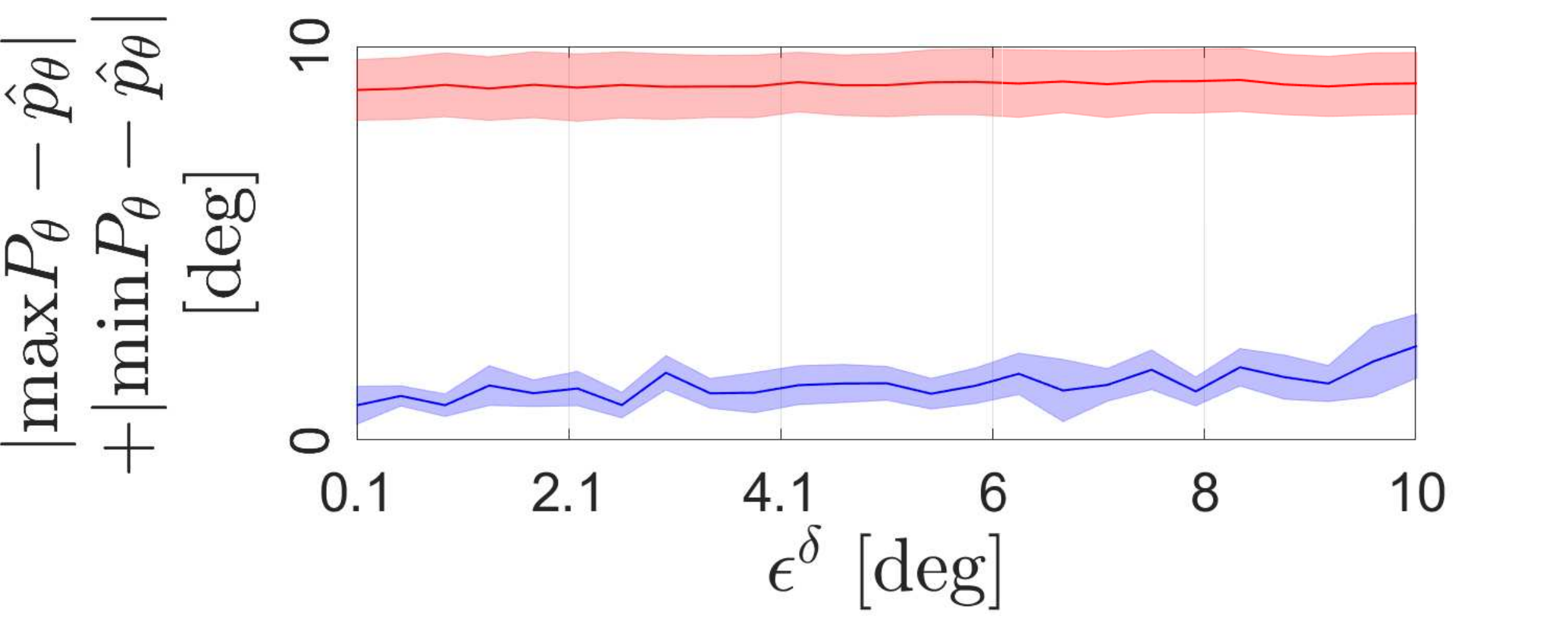}
        \caption{}
        \label{fig:sensitivityUnknown_Stereo_Delta_t}
    \end{subfigure}  
    &
\end{tabular}
\caption{Sensitivity analysis to control signal noise bounds: (a) vehicle body estimation with different velocity noise bounds $\epsilon^{v}$. (b) vehicle orientation estimation with different velocity noise bounds $\epsilon^{v}$. (c) vehicle body estimation with different steering noise bounds $\epsilon^{\delta}$. (d) vehicle orientation estimation with different steering noise bounds $\epsilon^{\delta}$.}
\label{fig:sensitivityCtrlParams}
\end{figure}

In this section, we conduct sensitivity analysis of the proposed algorithm to different sensor noise bounds $\epsilon^{w_a},\;\epsilon^{w_r}$, initialization uncertainties $V(P_{i}(0))$ and control signal noise bounds $\epsilon^{v}$, $\epsilon^{\delta}$, and compare the results with the ones using the FastSLAM. Same metrics used in Sec.~\ref{subsec:singleRun} are adopted here for performance comparison between different algorithms. The corresponding results are shown in Figs.~\ref{fig:sensitivityMeasurementParams},~\ref{fig:sensitivityInitializationParams},~\ref{fig:sensitivityCtrlParams}, respectively, where each data point on the solid line is the mean value of the aforementioned evaluation metrics in Sec.~\ref{subsec:singleRun} and the standard deviation is visualized using shaded strips.

As shown in Figs.~\ref{fig:sensitivityUnknown_Stereo_EWA_P} and~\ref{fig:sensitivityUnknown_Stereo_EWR_P}, the proposed algorithm is more robust to the changes in measurement noise bounds in vehicle body estimation, as the mean value of the metric $m_1$ stays at a steady level while the results from the FastSLAM fluctuate, and the standard deviation of the proposed method is smaller. Similar results can also be observed in vehicle orientation estimation as shown in Figs.~\ref{fig:sensitivityUnknown_Stereo_EWA_t} and~\ref{fig:sensitivityUnknown_Stereo_EWR_t}. In fact, the smaller standard deviations and less fluctuated mean values with the changing parameters are due to the fact that our method performs the set estimation in a deterministic way. In contrast, the FastSLAM re-samples particles at each time step, which increases the randomness and leads to larger standard deviation values. We also note that, as shown in Figs.~\ref{fig:sensitivityUnknown_Stereo_EWA_t} and~\ref{fig:sensitivityUnknown_Stereo_EWR_t}, the mean values from the proposed method are larger than the ones from the FastSLAM, which implies that the proposed method yields more conservative vehicle orientation estimates to guarantee $\hat{p}_{\theta}\in P_{\theta}$. 

The sensitivity analysis results to varying initial uncertainty set size of marker position are shown in Figs.~\ref{fig:sensitivityUnknown_Stereo_Pi_P} and~\ref{fig:sensitivityUnknown_Stereo_Pi_t}. The proposed method yields larger (smaller) mean values and smaller standard deviations in the vehicle body (orientation) estimates, which indicates that the proposed method estimates smaller vehicle body and orientation uncertainty sets compared to the ones by the FastSLAM. The sensitivity analysis results to varying control signal noise are shown in Figs.~\ref{fig:sensitivityUnknown_Stereo_V_P} and~\ref{fig:sensitivityUnknown_Stereo_Delta_P}. Though the two methods have similar performance in terms of the mean values of $m_1$, the proposed method can provide estimates with smaller standard deviation values. Again, the results of vehicle orientation estimation by the proposed method, as shown in Figs.~\ref{fig:sensitivityUnknown_Stereo_V_t} and~\ref{fig:sensitivityUnknown_Stereo_Delta_t}, are more conservative as a result of enforcing $\hat{p}_{\theta}\in P_{\theta}$. In conclusion, against uncertainties in the system parameters and initialization conditions, though the proposed algorithm has more conservative vehicle orientation estimates, it is more robust and ensures that the states are confined to the corresponding uncertainty sets, and it can provide smaller estimation error. For vehicle body estimates, the proposed method has a similar performance as the FastSLAM against measurement and control signal noise bounds, while being more robust to marker initialization uncertainties than the FastSLAM.

\subsection{Accommodating Sensor Uncertainties}\label{subsec:sensorUpdate}
\begin{figure}[t]
\centering
\includegraphics[width=0.8\linewidth]{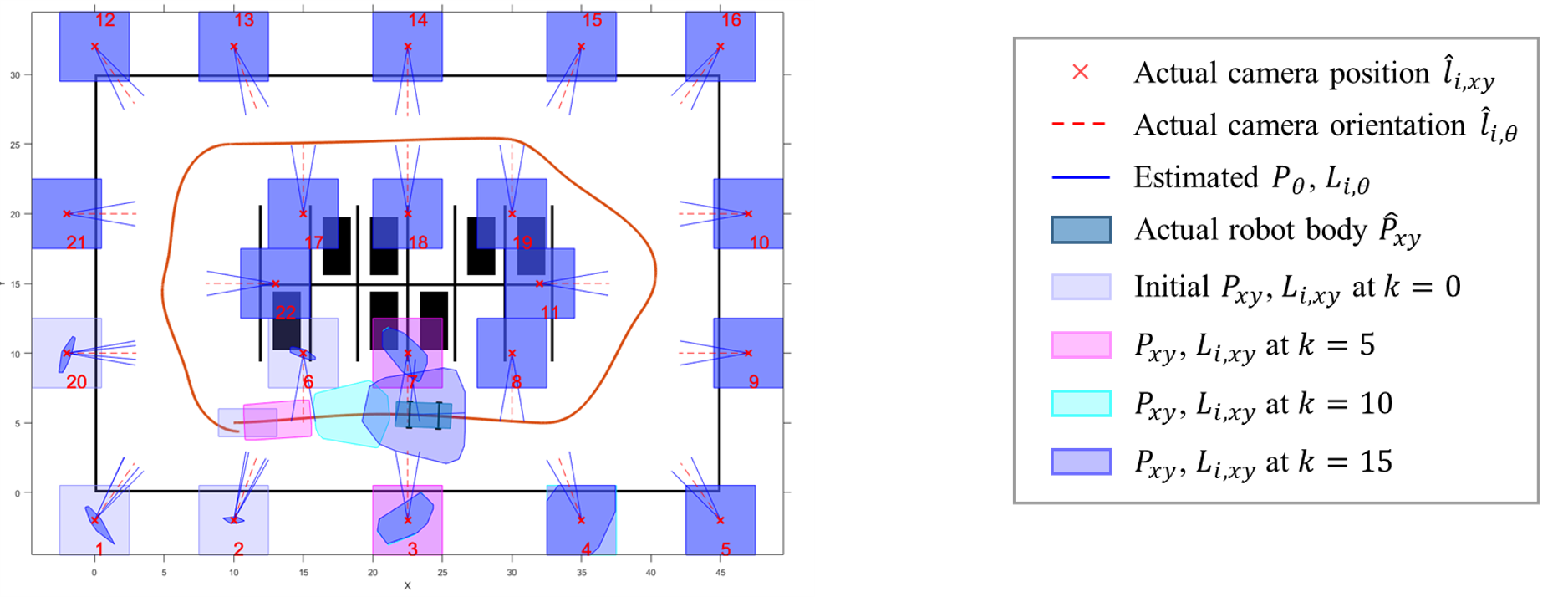}
\caption{Schematic of updating sensor uncertainty sets when the vehicle is accurately localized.}
\label{fig:sensorUpdate}
\end{figure}

At time step $k=0$, we initialize the proposed method with the same parameters as in Sec.~\ref{subsec:singleRun}. Meanwhile, we increase the uncertainties in the initial camera positions and orientations. As shown in Fig.~\ref{fig:sensorUpdate}, we initialize $\{L_{i,xy}\}_{i=1,\dots,m}$ as boxes of size $V(L_{i,xy})=25\;\rm m^2$, and $\{L_{i,\theta}\}_{i=1,\dots,m}$ as intervals of size $V(L_{i,\theta})=20\;\rm deg$. Fig.~\ref{fig:sensorUpdate} visualizes the sensors' and vehicle's uncertainty sets at four different simulation time steps, $k=0,5,10,15$. The sizes of the camera orientation and position uncertainty sets $L_{i,\theta},\;L_{i,xy},\; i=1,2,6,20$ are decreasing significantly as a result of the update process \eqref{eq:updateCamTheta},\eqref{eq:updateCamXY} with small uncertainties in vehicle orientation and body estimation at $k=0,5$. However, due to uncertainties in control signals, we observe enlarged vehicle body and orientation uncertainty sets from $k=0$ to $k=15$. Consequently, the updated uncertainty sets $L_{i,\theta},\;L_{i,xy},\; i=3,4,7$ have larger sizes than those of $L_{i,\theta},\;L_{i,xy},\; i=1,2,6,20$. Thus, a robot with small orientation and body uncertainty sets is able to mitigate the uncertainties in the sensor orientations and positions. This property can be applied in the sensor calibration process where the robot is well-localized using a third-party global positioning system allowing us to calibrate the sensor parameters and uncertainty sets using the robot localization information.

\subsection{Real-world Experiment Results}\label{subsec:lidarLoca}
\begin{figure}[t]
\centering
\includegraphics[width=0.8\linewidth]{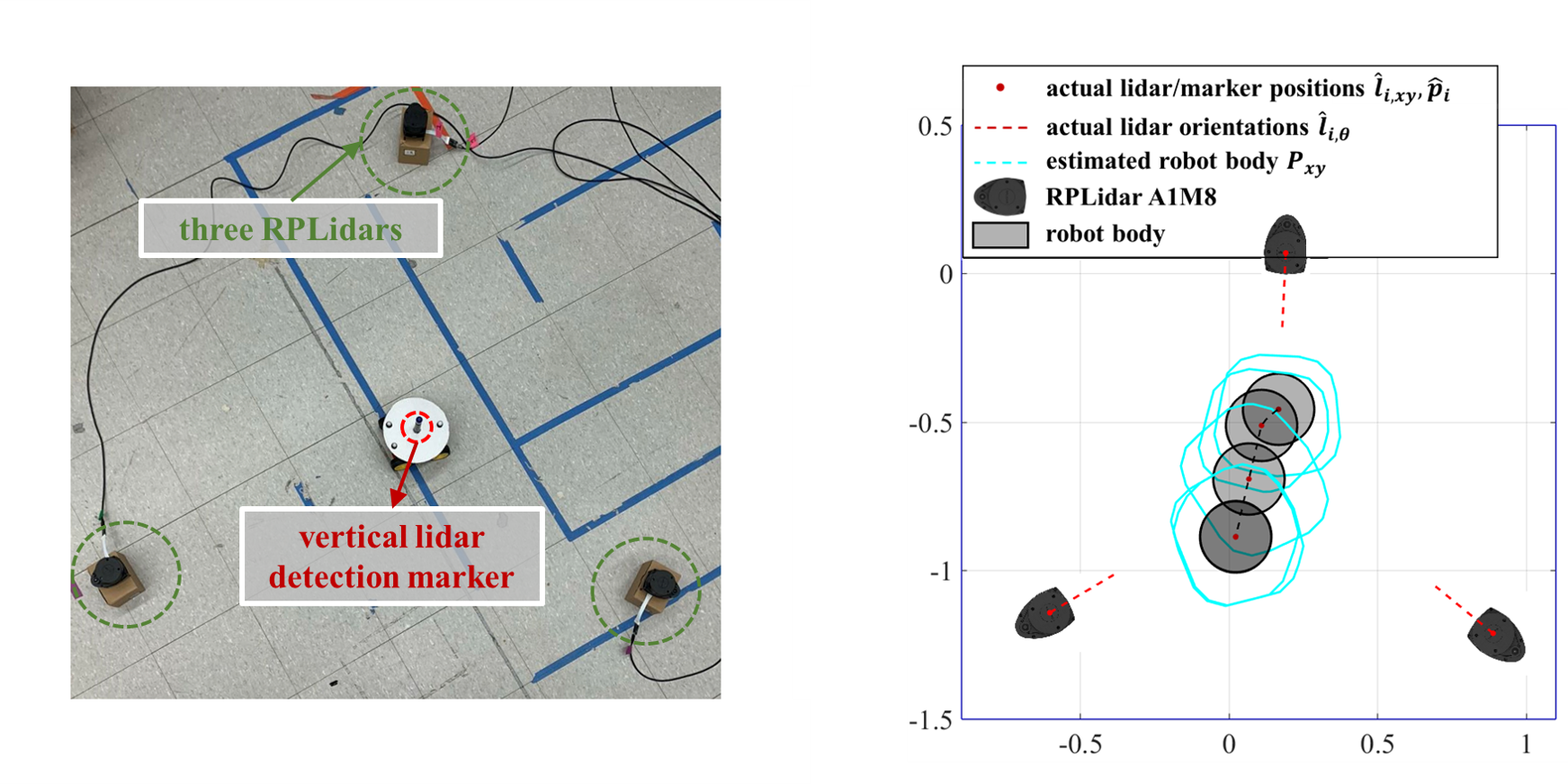}
\caption{Omnidirectional robot localization using lidar measurements: (Left) Photo of the test field with three RPLidars (infrastructure based sensors), and an omnidirectional robot attached with a detection marker. (Right) Test results using the proposed method.}
\label{fig:lidarLoca}
\end{figure}

To verify the applicability of the proposed method, we set up a sensing system to conduct real-time estimation of an omnidirectional robot body that is visualized as a circle of known radius $r$ in Fig.~\ref{fig:lidarLoca}. As shown in Fig.~\ref{fig:lidarLoca}, the localization system consists of three lidars (RPLidar A1M8~\cite{rplidar}). We attach a vertical bar to the center of the robot and use it as the lidar detection marker $\hat{p}_1$ such that the task is translated to estimate an uncertainty set $P_1$ that bounds the robot center, i.e., $\hat{p}_1\in P_1$. Subsequently, we can estimate the uncertainty set, i.e., $P_{xy} = P_1\oplus\mathcal{B}_2(r)$, that entirely bounds the robot body $\hat{P}_{xy}$ given the circle radius $r$. Meanwhile, to embed the omnidirectional robot kinematics into the algorithm setting, we define a maximum speed of the robot as $v_{max}$. Subsequently, we define $\epsilon^f=v_{max}\cdot dt$ and set $D_{i,x}(k)\times D_{i,y}(k)=\varnothing$ in the motion propagation~\eqref{eq:markerSetUpdate}. The measurement update can then be performed following the proposed method.
\footnotetext[2]{The python library is available at \url{https://pypi.org/project/pyrplidar/}}
\footnotetext[3]{The OptiTrack SDK is available at \url{https://optitrack.com/software/natnet-sdk/}}
\footnotetext[4]{A demontration video can be found at \url{https://user-images.githubusercontent.com/58400416/133294083-76bd6d9f-2807-4ab0-ba4e-ffa9abc69788.mp4}}

We attach three visual detection markers to the robot so that we can obtain the actual position of the robot center from the \textit{OptiTrack} motion capture system as the ground-truth information. The robot is controlled by user via Arduino platform where the maximum speed constraint is enforced. The lidar measurements are transmitted to a master computer, where the proposed localization is performed on MATLAB software, through USB connections and are decoded using 3rd party Python library~\footnotemark[2]. The OptiTrack measurements is transmitted to the master computer using a Python SDK~\footnotemark[3] provided by the OptiTrack. Based on the aforementioned kinematics assumption of the robot, we note that the synchronization of the control signal and lidar measurements can be conveniently achieved by setting $\epsilon^f=v_{max}\cdot dt$ where $dt=1/f$ and $f$ is the lidar measurement transmission frequency.

We first calibrate the range and angle measurement noise bounds as $\epsilon^{w_r}=0.073\;\rm m$ and $\epsilon^{w_a}=8.05^{\circ}$, respectively. With a robot trajectory that covers the majority area of the test field, we calibrate the noise bounds as the maximum errors between measurements from \textit{OptiTrack} and the ones from lidars. The robot navigates in the test field with a maximum speed of $v_{max}=0.10\;\rm m/s$. As shown in Fig.~\ref{fig:lidarLoca}, the proposed set theoretic localization method guarantees that the estimated uncertainty set (green line) always contains the robot body (circle with a radius of $0.12\;\rm m$). This result demonstrates the possibility for real-world implementation of the proposed method.~\footnotemark[4]
\section{Conclusion}\label{sec:conclusion}
In this paper, a set-theoretic localization algorithm that relies on the infrastructure-based sensing has been proposed. The theoretical properties and computational approaches for this set-theoretic localization method have been established. The theoretical properties have also been validated through simulations and real-world experiments. Sensitivity analysis to uncertainties in system parameters and initialization conditions has been conducted. By comparing with the state-of-the-art FastSLAM algorithm, the sensitivity analysis results demonstrated that the proposed method was more robust and ensured that the states were confined to the corresponding uncertainty sets, yet provided smaller estimation errors. Future work will focus on extending the proposed method to localization problems to a higher-dimensional state space, i.e., position and orientation estimation of aerial vehicles.
\appendix
\section{Marker Kinematics}\label{sec:appendix-kinematics}
\begin{figure}[h]
\centering
    \begin{subfigure}[c]{0.45\textwidth}
        \includegraphics[width=0.8\linewidth]{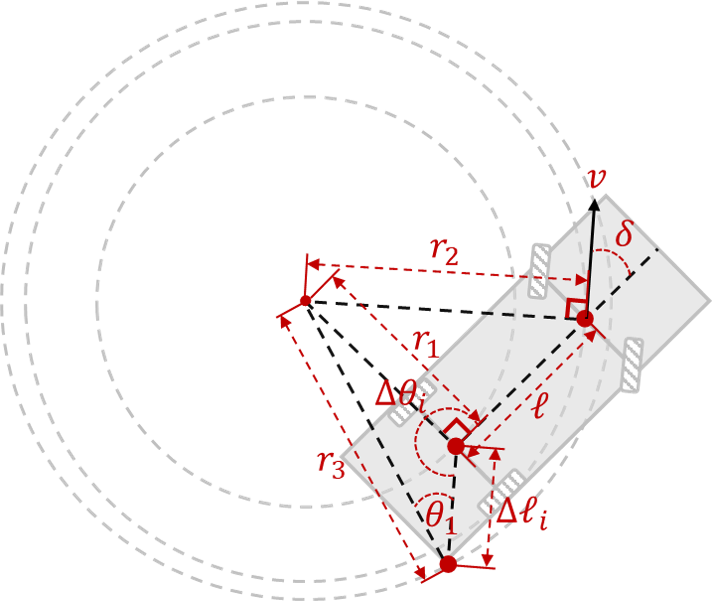}
        \caption{}
        \label{fig:kinematics_1}
    \end{subfigure}    
    \begin{subfigure}[c]{0.45\textwidth}
        \includegraphics[width=0.95\linewidth]{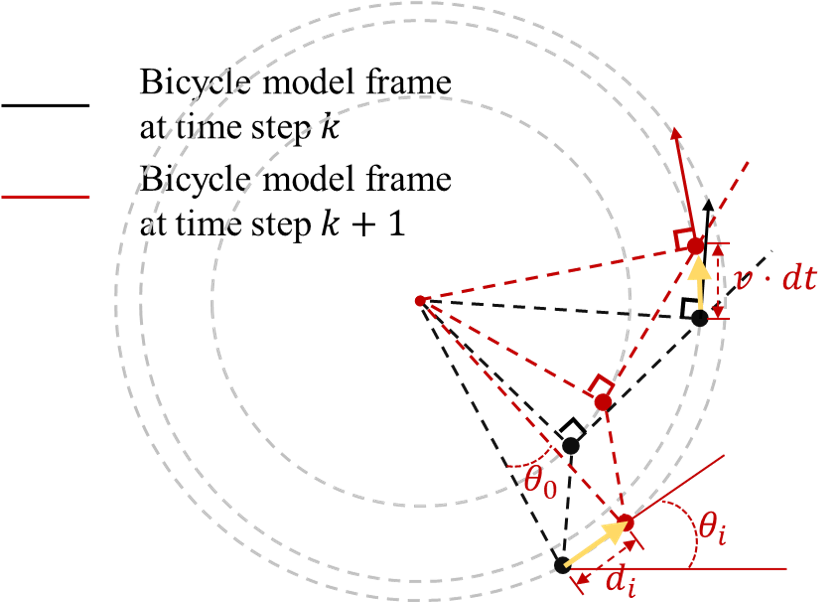}
        \caption{}
        \label{fig:kinematics_2}
    \end{subfigure} 
\caption{Illustration of the $i^{th}$ marker's kinematics.}
\label{fig:robot_kinematics}
\end{figure}

We use the bicycle model to simplify the kinematics of the vehicle as shown in Fig.~\ref{fig:kinematics_1}. Based on the geometric relationship, we can derive the following equations,
\begin{equation*}
    r_1 = \frac{\ell}{\tan \delta},\;r_2 = \frac{\ell}{\sin \delta},
\end{equation*}
whereby we can obtain the following quantities,
\begin{equation*}
    \theta_0 = \frac{v\cdot dt}{r_2},\;
    r_3 = \left( r_1^2 + \Delta\ell_i^2 - 2r_1\Delta\ell\cos(\Delta\theta_i-\frac{\pi}{2}) \right)^{1/2}.
\end{equation*}
Then, as shown in Fig.~\ref{fig:kinematics_2}, the displacement of the $i^{th}$ marker is equal to 
\begin{equation}\label{eq:kinematics_d}
    d_i(v,\delta) =\theta_0r_3= v\cdot dt\cdot \sqrt{(\frac{\Delta \ell_i\sin{\delta}}{\ell})^2+(\cos{\delta})^2-\frac{\Delta \ell_i}{\ell}\cdot\sin{\Delta \theta_i}\cdot\sin{(2\delta)}}.
\end{equation}
Moreover, it can be shown that 
\begin{equation*}
    \sin{\theta_1} = \frac{r_1}{\Delta\ell_i}\sin{(\frac{3\pi}{2}-\Delta\theta_i-\theta_1)}
\end{equation*}
which yields 
\begin{equation*}
\theta_1 = \atan2\left(-\ell\cos{\Delta \theta_i}, \Delta\ell_i\tan{\delta}-\ell\sin{\Delta \theta_i}\right).
\end{equation*}
Afterwards, the angle $\theta_i$ of the $i^{th}$ marker satisfies 
\begin{equation}\label{eq:kinematics_t}
    \theta_i(v,\delta, \hat{p}_{\theta}(k)) =  \hat{p}_{\theta}(k) + \Delta \theta_i +\theta_1 - \frac{3\pi}{2} = \hat{p}_{\theta}(k) + \Delta \theta_i + \atan2\left(\Delta\ell_i\tan{\delta}-\ell\sin{\Delta \theta_i}, \ell\cos{\Delta \theta_i}\right).
\end{equation}
The kinematics model in \eqref{eq:motionModel} can be shown from \eqref{eq:kinematics_d} and \eqref{eq:kinematics_t}.

\section{Proof of Proposition 1}\label{sec:appendix-proof}
We assume the following condition  
\begin{equation}\label{eq:nominalInUnSet}
\begin{gathered}
    \hat{l}_{i,xy}(k) \in L_{i,xy}(k),\; \hat{l}_{i,\theta}(k) \in L_{i,\theta}(k),\hat{p}_j(k) \in P_{j}(k),\\
    i=1,\dots,m,\; j=1,\dots,n,
\end{gathered}
\end{equation}
holds for $k=0$ during initialization. Through the propagation and update of uncertainty sets, if the condition in (\ref{eq:nominalInUnSet}) being true for $k+1$ can be induced from same condition being satisfied at time step $k$, the actual states $\hat{l}_{i,xy},\;\hat{l}_{i,\theta},\;\hat{p}_j$ stay in the corresponding uncertainty sets $ L_{i,xy},\; L_{i,\theta},\;P_{j}$, respectively, by principle of induction. Afterward, it's convenient to verify that the robot body and orientation are contained in $P_{xy},P_{\theta}$. For simplification, we provide the proof for the monocular camera that the stereo case resembles.
\subsection{Motion Propagation}\label{appendix:motionProp}
First, we examine the propagation process in (\ref{eq:markerSetUpdate}),~(\ref{eq:camSetUpdateDecomposed}). The displacement vector in \eqref{eq:motionModel} can be shown to satisfy the following property
\begin{equation*}
    \left[
    \begin{array}{c}
        \hat{dx}\\
        \hat{dy} 
    \end{array}
    \right]
    =
    \left[
    \begin{array}{c}
        \hat{p}_{i,x}(k+1)-\hat{p}_{i,x}(k)\\
        \hat{p}_{i,y}(k+1)-\hat{p}_{i,y}(k) 
    \end{array}
    \right]
    \in D_{i,xy}(k)
\end{equation*}
where 
\begin{equation*}
D_{i,xy}(k) =
\left\{
    \left[
    \begin{array}{c}
        dx\\
        dy
    \end{array}
    \right]
    \in \mathbb{R}^2
    :
    \left[
    \begin{array}{c}
        dx\\
        dy
    \end{array}
    \right]
    =
    \left[
    \begin{array}{c}
        d_i(v, \delta)\cdot\cos{(\theta_i(v,\delta,p_{\theta}))}\\
        d_i(v, \delta)\cdot\sin{(\theta_i(v,\delta,p_{\theta}))}
    \end{array}
    \right]
    ,\;\abs{v-\hat{v}}\leq\epsilon^{v},\;\abs{\delta-\hat{\delta}}\leq\epsilon^{\delta},\;p_{\theta}\in P_{\theta}(k)
\right\}.
\end{equation*}
By the definition of $D_{i,x}(k),D_{i,y}(k)$, we have $D_{i,xy}(k)\subset D_{i,x}(k)\times D_{i,y}(k)$. Considering $w_f^i\in \mathcal{B}_{\infty}(\epsilon^{f})$ and $\hat{p}_j(k)\in P_j(k)$, the set inclusion $\hat{p_j}(k+1|k) \in P_j(k+1|k)$ is guaranteed by the Minkowski sum in  (\ref{eq:markerSetUpdate}). Given that the sensors are stationary, $\hat{l}_{i,xy}(k+1|k) \in L_{i,xy}(k+1|k)$ and $\hat{l}_{i,\theta}(k+1|k) \in L_{i,\theta}(k+1|k)$ follow from (\ref{eq:nominalInUnSet}) given (\ref{eq:camSetUpdateDecomposed}).
\subsection{Measurement-to-marker Correspondence}
Consider measurements from the $i^{th}$ camera. We now show that the proposed method in Sec.~\ref{subsec:correspondance} can guarantee the possible solutions $C_{l_i}^{(\mu)},\;\mu=1,\dots,c$, contain the actual one $C_{l_i}^{(\mu^*)}$. Given $M_{l_i}=\{\alpha^{(q)}_{i,j^*}(k+1)\}$, we suppose the ${q}^{th}$ measurement $\alpha^{(q)}_{i,j^*}$ is, in fact, a measurement of the ${j^*}^{th}$ marker. In the reference frame taking $\hat{l}_{i,xy}$ as origin, the coordinates of the ${j^*}^{th}$ marker are $\hat{p}_{j^*}(k+1|k)-\hat{l}_{i,xy}(k+1|k)$ and satisfy
\begin{equation*}
    w_a^{i,j^*}(k+1) = \alpha^{(q)}_{i,j^*}(k+1) - \atan2(\hat{p}_{j^*,y}-\hat{l}_{i,y},\;\hat{p}_{j^*,x}-\hat{l}_{i,x}) + \hat{l}_{i,\theta},
\end{equation*}
by \eqref{eq:measurementModel} where we omit the notations $(k+1|k)$ for simplicity. Since the noise $\abs{w_a^{i,j^*}(k+1)}\leq \epsilon^{w_a}$ is bounded and $\hat{l}_{i,\theta}(k+1|k)\in L_{i,\theta}(k+1|k) \subset [\theta_0-\delta \theta,\theta_0+\delta \theta]$, it can be shown from \eqref{eq:PMSetMatching} that
\begin{equation*}
    \hat{p}_{j^*}(k+1|k)-\hat{l}_{i,xy}(k+1|k) \in P'_M(l_i,\alpha_{i,j^*}^{(q)}).
\end{equation*}
Given $\hat{l}_{i,xy}(k+1|k)\in L_{i,xy}(k+1|k)$ from \ref{appendix:motionProp} and equation above, the actual marker position satisfies
\begin{equation*}
    \begin{aligned}
    \hat{p}_{j^*}(k+1|k) &=  \hat{l}_{i,xy}(k+1|k) + \left(\hat{p}_{j^*}(k+1|k)-\hat{l}_{i,xy}(k+1|k) \right)\\
    &\in L_{i,xy}(k+1|k) \oplus P'_{M}(l_i, \alpha^{(q)}_{i,j^*}).
    \end{aligned}
\end{equation*}

Furthermore, as $\hat{p}_{j^*}(k+1|k)\in P_{j^*}(k+1|k)$ from \ref{appendix:motionProp}, the nonempty condition in \eqref{eq:nonemptyMatchingCondition} is satisfied, therefore, $C_{l_i}(q, j^*)=1$. Same conclusion can be applied to each single measurement in $M_{l_i}(k+1)$. Thus, the actual solution $C_{l_i}^{(\mu^*)}$ is guaranteed to be presented in the correspondence matrix $C_{l_i}(k+1)$. An algorithm examining the logical self-consistency in $C_{l_i}(k+1)$ will keep $C^{(\mu^*)}_{l_i}$ as one of the solutions $C^{(\mu)}_{l_i}$, $\mu \in \{1,\dots,c\}$.

\subsection{Measurement Update}
In the following discussion, we first present the proof with only one measurement-to-marker solution that is actual, then, the extension to case with multiple correspondences will be presented. 
\subsubsection{Update of \texorpdfstring{$L_{i,\theta}$}{Lg}}\label{appendix:GuaranteedCamTheta}
Given only one solution of $M_{l_i}(k+1)=\{\alpha^{(q)}_{i,j}(k+1)\}$ that is actual, we can obtain $\hat{l}_{i,\theta}(k+1)\in [\psi_{i,j}, \phi_{i,j}]$ for each measurement $\alpha^{(q)}_{i,j}(k+1)$ by \eqref{eq:measurementModel} and \eqref{eq:betaMinMax}. Indeed, consider the entire set of measurements, we have
\begin{equation*}
    \hat{l}_{i,\theta}(k+1)\in \bigcap\limits_{\alpha^{(q)}_{i,j}\in M_{l_i}}[\psi_{i,j}, \phi_{i,j}],
\end{equation*}
which, combining with $\hat{l}_{i,\theta}(k+1) \in L_{i,\theta}(k+1|k)$, proves \eqref{eq:updateCamTheta_KnownMatching}. Moreover, if multiple solutions $C^{(\mu)}_{l_i},\;\mu=1,\dots,c$ are given where the actual one $C^{(\mu^*)}_{l_i}$ is contained, the following relationship can also be established
\begin{equation*}
    \hat{l}_{i,\theta}(k+1)\in \bigcap\limits_{\substack{q,\; \alpha^{(q)}_{i,j'}\in M_{l_i},\\C^{(\mu^*)}_{l_i}(q,j)=1}}[\psi_{i,j}, \phi_{i,j}]\subset \bigcup\limits_{\mu=1}^{c}\left(\bigcap\limits_{\substack{q,\; \alpha^{(q)}_{i,j'}\in M_{l_i},\\C^{(\mu)}_{l_i}(q,j)=1}}[\psi(L_{i,xy}, P_{j}, \alpha^{(q)}_{i,j'}), \phi(L_{i,xy}, P_{j}, \alpha^{(q)}_{i,j'})]\right).
\end{equation*}
Given $\hat{l}_{i,\theta}(k+1)\in L_{i,\theta}(k+1|k)$ and equation above, we can show $\hat{l}_{i,\theta}(k+1)\in L_{i,\theta}(k+1)$ as in \eqref{eq:updateCamTheta}.
\subsubsection{Update of \texorpdfstring{$L_{i,xy}$}{Lg}}\label{appendix:GuaranteedCamXY}
Given measurement $\alpha_{i,j}$ and in a reference frame centered at $\hat{p}_j$, the coordinates of the $i^{th}$ camera $[l'_x\;l'_y]^T=\hat{l}_{i,xy}(k+1|k)-\hat{p}_j(k+1|k)$ satisfy
\begin{equation*}
    \atan2(-l'_y,-l'_x) = \alpha_{i,j} + \hat{l}_{i,\theta}(k+1|k) - w_a^{i,j}(k+1)
\end{equation*}
by  (\ref{eq:measurementModel}). Since the noise $\abs{w_a^{i,j}(k+1)}\leq \epsilon^{w_a}$ and $\hat{l}_{i,\theta}(k+1|k)\in L_{i,\theta}(k+1|k) \subset [\theta_c -\delta \theta_c,\theta_c+\delta\theta_c]$, it can be shown that 
\begin{equation*}
    \hat{l}_{i,xy}(k+1|k)-\hat{p}_{j}(k+1|k) \in  L_{M}(p_{j}, \alpha_{i,j})
\end{equation*}
from  (\ref{eq:measurementSetLocalAtMarker}). Given $\hat{p}_{j}(k+1|k)\in P_{j}(k+1|k)$ from propagation and equation above, the actual camera  position satisfies 
\begin{equation*}
    \begin{aligned}
    \hat{l}_i(k+1|k) &=  \hat{p}_j(k+1|k) + \left(\hat{l}_{i,xy}(k+1|k)-\hat{p}_j(k+1|k) \right)\\
    &\in P_j(k+1|k) \oplus L_{M}(p_{j}, \alpha_{i,j}).
    \end{aligned}
\end{equation*}
Considering all the measurements, the actual camera position is within the set intersection
\begin{equation*}
    \bigcap\limits_{\alpha_{i,j}\in M_{l_i}}\left( P_{j}(k+1|k) \oplus L_{M}(p_j, \alpha_{i,j})\right),
\end{equation*}
which together with $\hat{l}_{i,xy}(k+1)\in L_{i,xy}(k+1|k)$ proves \eqref{eq:updateCamXY_KnownMatching}.

Similar to \ref{appendix:GuaranteedCamTheta}, if multiple solutions are given including the actual one, the following relationship can also be established
\begin{equation*}
    \begin{aligned}
    \hat{l}_{i,xy}(k+1|k) &\in  \bigcap\limits_{\substack{q,\; \alpha^{(q)}_{i,j'}\in M_{l_i},\\C^{(\mu^*)}_{l_i}(q,j)=1}}\left( P_{j}(k+1|k) \oplus L_{M}(p_{j}, \alpha^{(q)}_{i,j'})\right)\\
    &\subset \bigcup\limits_{\mu=1}^{c}\left(\bigcap\limits_{\substack{q,\; \alpha^{(q)}_{i,j'}\in M_{l_i},\\C^{(\mu)}_{l_i}(q,j)=1}}\left( P_{j}(k+1|k) \oplus L_{M}(p_{j}, \alpha^{(q)}_{i,j'})\right)\right).
    \end{aligned}
\end{equation*}
Eventually, with $\hat{l}_{i,xy}(k+1)\in L_{i,xy}(k+1|k)$, we can prove $\hat{l}_{i,xy}(k+1)\in L_{i,xy}(k+1)$ as in \eqref{eq:updateCamXY}. 
\subsubsection{Update of \texorpdfstring{$P_j$}{Lg}}
It's likely that only a subset of cameras have corresponding measurements of the $j^{th}$ marker. Similarly, we note that 
\begin{equation*}
    \hat{p}_j(k+1) - \hat{l}_{i,xy}(k+1) \in  P_{M}(l_i, \alpha_{i,j}).
\end{equation*}
As $\hat{l}_{i,xy}(k+1)  \in L_{i,xy}(k+1)$ from \ref{appendix:GuaranteedCamXY}, we have 
\begin{equation*}
    \hat{p}_j(k+1) \in L_{i,xy}(k+1) \oplus P_{M}(l_i, \alpha_{i,j}).
\end{equation*}
Considering $M_{l_i}$, $i=1,\dots,m$ that contain measurement of the $j^{th}$ marker and $\hat{p}_j(k+1) \in P_{j}(k+1|k)$ from \ref{appendix:motionProp}, the actual marker position satisfies $\hat{p}_j(k+1) \in P_{j}(k+1)$ as derived in \eqref{eq:updateMarkerXY_KnownMatching}.

Again, given multiple solutions for $M_{l_i}$, one can only conclude $M_{l_i}$ certainly contains measurements of the $j^{th}$ marker if there is a corresponding measurement in all solutions, i.e., $\forall\mu\in\{1,\dots,c\},\; \exists q\leq n,\; C_{l_i}^{(\mu)}(q, j)=1$. Given such a set of measurements $M_{l_i}\in M^j$, we notice that 
\begin{equation*}
    \hat{p}_{j}(k+1) - \hat{l}_{i,xy}(k+1) \in P_{M}(l_i, \alpha_{i,j}) \subset \bigcup\limits_{\substack{\mu=1,\dots,c,\\\alpha^{(q)}_{i,j'}\in M_{l_i},\;C^{(\mu)}_{l_i}(q,j)=1}} P_{M}(l_i, \alpha^{(q)}_{i,j'}) .
\end{equation*}
Given $\hat{l}_{i,xy}(k+1)  \in L_{i,xy}(k+1)$ from \ref{appendix:GuaranteedCamXY}, we have 
\begin{equation*}
    \hat{p}_{j}(k+1) \in L_{i,xy}(k+1) \oplus \bigcup\limits_{\substack{\mu=1,\dots,c,\\\alpha^{(q)}_{i,j'}\in M_{l_i},\;C^{(\mu)}_{l_i}(q,j)=1}} P_{M}(l_i, \alpha^{(q)}_{i,j'}).
\end{equation*}
Consider all $M_{l_i}\in M^j,\;i=1,\dots,m$ and $\hat{p}_{j}(k+1) \in P_{j}(k+1|k)$, the actual marker position $\hat{p}_{j}(k+1) \in P_{j}(k+1)$ as derived in \eqref{eq:updateMarkerXY}.
\subsection{Robot Body and Orientation Estimation}
Points that are at most $r$ distance away from $p_i$ locate in $P_i\oplus\mathcal{B}_2(r)$. If $\norm{\hat{p}_i-\hat{p}_j}_2=r_{ij}$, the actual marker position $\hat{p}_j\in P_i\oplus\mathcal{B}_2(r_{ij})$. Thereby, the set refinement by rigid body constrains in (\ref{eq:rigidBodyRefine}) preserves the property that $\hat{p}_j\in P_j(k+1)$. Finally, given the assumption of the robot body being in the convex hull of the markers, $\hat{P}_{xy}$ in (\ref{eq:positionSetRecon}) over-bounds the entire robot body $P_{xy}$, i.e., $\hat{P}_{xy}\subset P_{xy}$. The proof of the actual robot orientation $\hat{p}_{\theta}\in P_{\theta}$ follows a similar procedure as in \ref{appendix:GuaranteedCamTheta}.
\bibliography{wileyNJD-AMA.bib}
\end{document}